\pdfoutput=1
\documentclass[11pt,a4paper]{article}

\usepackage[square,numbers,sort&compress]{natbib}

\usepackage{amsmath,amssymb}
\usepackage{mathtools}
\usepackage{bm}

\usepackage{graphicx}
\usepackage{float}
\usepackage{pdflscape}
\usepackage{caption}
\usepackage{subcaption}
\usepackage{booktabs}
\usepackage{longtable}
\usepackage{multirow}
\usepackage{array}
\usepackage{threeparttable}

\usepackage{xcolor}
\usepackage{colortbl}

\usepackage{fancyvrb}
\usepackage{tcolorbox}

\usepackage{enumitem}

\usepackage{algorithm}
\usepackage{algpseudocode}

\usepackage{hyperref}
\usepackage{bookmark}

\usepackage{cleveref}

\definecolor{codeblue}{RGB}{0,80,180}

\newcommand{\mn}[1]{{\small\texttt{#1}}}   \newcommand{\dn}[1]{{\small\texttt{#1}}}          

\DeclareRobustCommand{\_}{\textunderscore\allowbreak}
\newtcolorbox{findingbox}[1][]{
    colback=blue!5!white,
    colframe=blue!75!black,
    fonttitle=\bfseries,
    title=#1
}

\newtcolorbox{caveatbox}[1][]{
    colback=orange!5!white,
    colframe=orange!75!black,
    fonttitle=\bfseries,
    title=#1
}

\usepackage[T1]{fontenc}
\usepackage{tgpagella}
\usepackage{mathpazo}
\usepackage{inconsolata}
\usepackage[stretch=10,shrink=10]{microtype}

\usepackage[margin=1in]{geometry}
\usepackage{setspace}
\setlist{topsep=1ex, itemsep=0.8ex, parsep=0.3ex}

\usepackage{titlesec}
\titlespacing{\paragraph}{0pt}{1.5ex plus 0.5ex}{0.5em}

\usepackage{authblk}
\usepackage{orcidlink}

\hypersetup{
    colorlinks=true,
    linkcolor=codeblue,
    citecolor=codeblue,
    urlcolor=codeblue,
    pdfusetitle,
    hypertexnames=false,
}

\title{\textbf{Revisiting Chebyshev Polynomial and \\
        Anisotropic RBF Models for Tabular Regression}}

\author[1]{Luciano Gerber\,\orcidlink{0000-0002-8423-4642}\thanks{\texttt{L.Gerber@mmu.ac.uk}}}
\author[1]{Huw Lloyd\thanks{Now retired.}}

\affil[1]{Department of Computing and Mathematics,
    Manchester Metropolitan University,\protect\\
    Dalton Building, Chester Street,
    Manchester M1 5GD, United Kingdom}

\date{\today}
 
\graphicspath{{figures/}}

\begin{document}

\maketitle

\begin{abstract}
Smooth-basis models such as Chebyshev polynomial regressors and radial basis function (RBF) networks are well established in numerical analysis. Their continuously differentiable prediction surfaces suit surrogate optimisation and sensitivity analysis. Despite these properties, smooth models seldom appear in tabular regression, where tree ensembles dominate. We ask whether they can compete, benchmarking models across 55 regression datasets organised by application domain.

We develop an anisotropic RBF network with data-driven centre placement and gradient-based width optimisation, a ridge-regularised Chebyshev polynomial regressor, and a smooth-tree hybrid (Chebyshev model tree), with all three released as scikit-learn-compatible packages. We benchmark these against tree ensembles, a bin-based additive model, a pre-trained transformer and standard baselines, evaluating accuracy alongside the train--test gap and prediction stability under retraining.

The transformer ranks first on accuracy across a majority of datasets, but its GPU dependence, prediction cost, and dataset-size limits constrain deployment in CPU-based settings. Among CPU-viable models, smooth models and tree ensembles are statistically tied on accuracy at the medium data scale that characterises most of our datasets ($n \lesssim 50\,000$, $d \lesssim 50$). In a full-scale check on eight datasets, XGBoost achieved the highest mean accuracy, with the Chebyshev model tree close behind. Where accuracy is matched, smooth-basis models tend to exhibit tighter train--test gaps. Across the benchmark, the gap is positively associated with prediction stability under retraining, supporting its inclusion alongside predictive accuracy. We recommend including smooth-basis models among the candidate models for tabular regression, particularly when downstream use benefits from smooth prediction surfaces.
 \end{abstract}


\section{Introduction}
\label{sec:intro}

Practitioners choosing a regression model for tabular problems tend to reach for tree ensembles (e.g., random forests, gradient-boosted trees), which have dominated benchmark leaderboards on predictive accuracy \citep{grinsztajn2022, fernandez-delgado2014}.
This paper examines whether two smooth-basis models from numerical analysis, Chebyshev polynomial regressors and radial basis function (RBF) networks, can match tree ensembles on predictive accuracy while offering complementary strengths in the train--test gap, smoothness, and interpretability.
Both are well established in function approximation but seldom applied to tabular regression, despite offering practical advantages alongside accuracy.
They are directly usable as surrogates in gradient-based optimisation \citep{forrester2009recent,jones1998efficient}, sensitivity analysis, and other settings where predictions must vary gradually with inputs.
Moreover, Chebyshev models yield explicit polynomial coefficients that can be inspected or constrained \citep{trefethen2019approximation}, while RBF models are weighted sums of localised basis functions whose centres and widths carry geometric meaning \citep{buhmann2003rbf}.
These models also train and predict on commodity hardware without GPU acceleration, a practical consideration for the many applied-science and engineering settings where specialised compute is unavailable or uneconomical.

Tabular benchmarks primarily report predictive accuracy \citep{shwartz2022tabular, mcelfresh2023neural}.
We also report the train--test gap, defined as training $R^2$ minus test $R^2$.
Stability theory links sensitivity to changes in the training sample with bounds on the generalisation gap \citep{bousquet2002stability}, motivating our empirical comparison of the train--test gap with prediction stability under retraining.

We revisited these smooth-basis models under contemporary benchmarking standards and with modern optimisation techniques.
For RBF networks, we developed an anisotropic formulation (\mn{erbf}) with data-driven centre placement, width initialisation, and gradient-based optimisation of per-dimension widths. These design choices aim to mitigate the instabilities of traditional simultaneous centre-and-width optimisation.
For Chebyshev regressors (\mn{chebypoly}), we combined optional pairwise interaction terms with ridge regularisation, synthesising known numerical-analysis building blocks into a practical tool for tabular regression.
For problems where the response may contain discontinuities, we also evaluated Chebyshev model trees (\mn{chebytree}), which partition the input space using tree splits and fit a Chebyshev polynomial in each leaf.
We benchmarked them against established baselines (ridge regression, \mn{ridge}, and a single decision tree, \mn{dt}, representing opposite ends of the bias-variance spectrum), an Explainable Boosting Machine (\mn{ebm}) \citep{nori2019interpretml} as a representative of additive, bin-based models, tree ensembles (Random Forest, \mn{rf}, and XGBoost, \mn{xgb}), and a pre-trained tabular foundation model (TabPFN, \mn{tabpfn}) \citep{hollmann2025tabpfn}\footnote{Our experiments use TabPFN v2.5 weights (\texttt{tabpfn}~6.3.1), distributed under a non-commercial licence.} as a representative of the rapidly developing family of transformer-based tabular learners.
The comparison spans 55 regression datasets from four domain strata (engineering and simulation; behavioural and social; physics, chemistry, and life sciences; economics and pricing).

\paragraph{Contributions}

\begin{enumerate}[leftmargin=*, itemsep=0.5ex]
\item \textbf{Multi-axis benchmark.}
We benchmarked nine models across 55 datasets on predictive accuracy, the train--test gap and computational cost. We used bootstrap retraining to test how well the gap serves as an inexpensive proxy for prediction stability.

\item \textbf{Model implementations.}
The primary modelling contribution is \mn{erbf}, an anisotropic RBF network with a three-stage training pipeline: data-driven centre placement, width initialisation, and decoupled gradient-based width optimisation in log-space.
We also provide a Chebyshev polynomial regressor (\mn{chebypoly}) and a Chebyshev model tree (\mn{chebytree}); these synthesise established numerical-analysis components into practical tabular regression tools.
All three are sklearn-compatible estimators, available on PyPI (\texttt{erbf}, \texttt{poly-basis-ml}).
\end{enumerate}

\paragraph{Summary of findings}
TabPFN ranks first on accuracy across a majority of datasets but requires GPU inference, is constrained by dataset size, and incurs high prediction cost.
Among CPU-viable models, the six competitive models (\mn{erbf}, \mn{chebytree}, \mn{xgb}, \mn{chebypoly}, \mn{ebm}, \mn{rf}) are statistically indistinguishable on accuracy (Friedman test with Nemenyi post-hoc comparison, $\alpha\!=\!0.05$), but the smooth-basis models (\mn{erbf}, \mn{chebypoly}) show the smaller train--test gap in 87\% of pairwise comparisons at matched accuracy ($|\Delta\bar{R}^2| \leq 0.02$).
We recommend including smooth-basis models among the candidate models for tabular regression, particularly when downstream use benefits from smooth prediction surfaces.

The remainder of this paper is organised as follows.
\Cref{sec:background} reviews the model families, their theoretical foundations, and positions the work relative to existing tabular benchmarks and alternative approaches.
\Cref{sec:methods} describes the model architectures, benchmark design, and evaluation protocol.
\Cref{sec:results} presents results on accuracy, the train--test gap, and computational cost.
\Cref{sec:taxonomy} develops a practical model selection framework.
\Cref{sec:discussion} discusses limitations and implications, and \Cref{sec:conclusion} concludes.
 \section{Background and Related Work}
\label{sec:background}

\subsection{Radial basis function networks}
\label{sec:bg-rbf}

Radial basis function (RBF) networks approximate a target function as a weighted sum of localised kernels centred at selected points in the input space \citep{broomhead1988radial, buhmann2003rbf}.  Centre selection strategies range from random subsets and $k$-means clustering to greedy procedures that maximise explained variance \citep{chen1991ols}.

Classical RBF networks have seen limited adoption in applied machine learning.  Two commonly cited difficulties are that simultaneously optimising centres and widths is non-convex, and that isotropic widths (a single width per basis function) adapt poorly to anisotropic structure in the data \citep{buhmann2003rbf, zhou2019hbfn}.

The generalisation from isotropic to anisotropic basis functions has been explored under several names.  \citet{zhou2019hbfn} provide a comprehensive survey of \emph{hyper basis function} neural networks (HBFNNs), which replace the Euclidean norm with a weighted or Mahalanobis norm, allowing each unit to stretch along different axes.  Our \mn{erbf} model adopts diagonal per-dimension widths (a special case of the HBFNN framework) but contributes a decoupled three-stage training pipeline that sidesteps the joint optimisation difficulties described above; the full design is described in \cref{sec:methods}.

\subsection{Chebyshev polynomials and polynomial regression}
\label{sec:bg-chebyshev}

Chebyshev polynomials of the first kind form an orthogonal basis under the Chebyshev weight on $[-1,1]$, with numerical properties that make them attractive for regression: they can improve conditioning relative to monomials at the same degree, although the conditioning of the design matrix depends on the sampled inputs \citep{trefethen2019approximation}.

Despite this strong theoretical foundation, Chebyshev polynomials have seldom been used as a basis for supervised regression in machine learning.  Two well-established alternatives that emphasise smoothness or interpretability are piecewise or additive in structure: MARS \citep{friedman1991mars} fits adaptive hinge-function splines and generalised additive models \citep{hastie1990gam} sum univariate smooth functions.  These methods enforce smoothness within each component or segment but not globally across the prediction surface.  Our approach combines a Chebyshev expansion with ridge regularisation \citep{hoerl1970ridge} to provide a globally smooth alternative: the formulation is linear in its parameters once the basis is constructed, so fitting reduces to a single regularised least-squares solve, yielding a smooth, globally defined predictor.  The \mn{chebypoly} implementation is described in \cref{sec:methods}.

\subsection{Model trees}
\label{sec:bg-modeltrees}

Model trees combine recursive partitioning with local regression models at the leaves.  \citet{quinlan1992learning} introduced M5, which fits linear models in each leaf node; \citet{wang1997modeltrees} refined the approach as M5$'$ with simplified pruning.  The appeal of model trees lies in their ability to capture regime changes through the tree structure while modelling smooth relationships within each regime.  Our \mn{chebytree} extends this idea by replacing linear leaf models with low-degree Chebyshev polynomial fits (\cref{sec:methods}).

\subsection{Smoothness, stability, and generalisation}
\label{sec:bg-smoothness}

Smoothness serves as an implicit regulariser across much of statistical learning.  \citet{poggio1990networks} showed that when function approximation is cast as a variational problem penalising derivatives of the fitted surface, the optimal solution takes the form of an RBF or kernel expansion, and ridge regression \citep{hoerl1970ridge} promotes smoother fits by penalising coefficient magnitude.  In a foundational result, \citet{bousquet2002stability} linked algorithmic stability -- insensitivity to perturbations of individual training examples -- to generalisation, bounding leave-one-out error for sufficiently stable learners.  We invoke this framework as motivation rather than as a formal guarantee: verifying the Lipschitz conditions it requires for each learning algorithm lies beyond the scope of an empirical benchmark.  \citet{belkin2019reconciling} demonstrated that this classical picture requires revision in the interpolation regime, where over-parameterised models can generalise well despite fitting training data exactly, but for the moderately parameterised models in our benchmark, the stability-generalisation link remains directly applicable.

With these foundations in place, we now review the broader landscape of tabular regression benchmarks and alternative model families that motivate our experimental design.

\subsection{Tabular benchmarks and evaluation}
Influential large-scale comparisons consistently rank tree ensembles (random forests, \citealp{breiman2001random}; gradient-boosted trees, \citealp{chen2016xgboost}) as the strongest general-purpose learners for tabular data \citep{fernandez-delgado2014,grinsztajn2022,shwartz2022tabular}.  Recent work adds nuance: \citet{mcelfresh2023neural} show that neural architectures occasionally surpass boosted trees when hyperparameters are tuned on equal footing, and TabPFN \citep{hollmann2025tabpfn} achieves state-of-the-art accuracy on small-to-moderate datasets through a pre-trained transformer.  On the methodological side, \citet{demsar2006statistical} established the widely adopted Friedman and Nemenyi tests for multi-classifier comparison, and \citet{cawley2010over} demonstrated the necessity of nested cross-validation to avoid selection bias. TabArena, the evaluation benchmark introduced alongside TabPFN~2.5 \citep{grinsztajn2025tabpfn}, extends this tradition to industry-scale datasets with up to 100\,000 training samples, though its evaluation focuses on accuracy rankings.  Our benchmark adopts nested cross-validation with Optuna-based inner tuning \citep{akiba2019optuna} and reports the train--test gap alongside accuracy, an axis on which model families diverge even when accuracy is matched.

Several recent model families share our interest in smooth, interpretable, or non-tree alternatives for tabular data, each emphasising different trade-offs.

\subsection{Additive and shape-constrained models}
\label{sec:bg-additive}
Explainable Boosting Machines (EBMs)~\citep{nori2019interpretml} use boosting to estimate additive terms on binned features, while Neural Additive Models (NAMs)~\citep{agarwal2021nam} use per-feature neural networks. We include EBM as a bin-based additive comparator.
Kolmogorov--Arnold Networks (KANs)~\citep{liu2025kan} place learnable spline activations on the edges of a network.
In our benchmark, \mn{chebypoly} augments its per-feature polynomial basis with optional pairwise interaction terms, while \mn{erbf}'s multivariate Gaussian kernels couple the input dimensions within each basis function.

\subsection{Kernel methods and Gaussian processes}
Kernel ridge regression and support vector regression scale as $O(n^{3})$ or $O(n^{2})$ in time and memory without approximation, making them impractical for many of the datasets in our benchmark.  Gaussian processes \citep{rasmussen2006gp} provide principled smoothness control through the kernel length-scale and deliver calibrated uncertainty estimates, but suffer the same cubic scaling; inducing-point methods \citep{titsias2009variational} alleviate the cost but introduce additional hyperparameters and approximation error whose impact varies across datasets.  Our \mn{erbf} sidesteps this cost structure by fixing $K$ centres ($K \ll n$) and fitting per-centre widths via analytical gradients, yielding $O(nKd)$ complexity with no kernel matrix inversion.

\subsection{Deep learning for tabular data}
FT-Transformer \citep{gorishniy2021revisiting}, TabNet \citep{arik2021tabnet}, and NODE \citep{popov2019node} adapt attention, gating, and differentiable oblivious decision trees, respectively, to tabular inputs.  These architectures can match or exceed tree ensembles on sufficiently large datasets; they occupy a different part of the design space from our models, trading smoothness guarantees and CPU-based training for greater representational flexibility.  TabPFN \citep{hollmann2025tabpfn} takes a distinct approach: pre-trained on millions of synthetic datasets drawn from a learned prior over data-generating processes, it performs in-context learning at inference time without task-specific training or hyperparameter tuning, achieving strong accuracy on small-to-moderate datasets.  We include it in our benchmark as a neural comparator.  The field continues to evolve rapidly: tabular foundation models such as TabICL \citep{qu2025tabicl} and TabPFN~2.5 \citep{grinsztajn2025tabpfn} now match or exceed tree ensembles on classification benchmarks.
 \section{Methods}
\label{sec:methods}

This section describes the three models we contribute to the benchmark: an anisotropic RBF network (\mn{erbf}), a Chebyshev polynomial regressor (\mn{chebypoly}), and a hybrid Chebyshev model tree (\mn{chebytree}).  \mn{erbf} and \mn{chebypoly} produce globally smooth (continuously differentiable) prediction surfaces; \mn{chebytree} produces piecewise-smooth surfaces, with smooth polynomial fits within each tree leaf.
For each, we outline the mathematical formulation and the algorithmic refinements that make it competitive on tabular regression.
We also describe the Explainable Boosting Machine (\mn{ebm}), included as an additive, bin-based comparator.
Benchmark design and the evaluation protocol are deferred to \cref{sec:benchmark-design}.
Software packages, API details, and installation instructions are given in \cref{sec:software-details,sec:installation}.

\subsection{Anisotropic RBF Network (\mn{erbf})}
\label{sec:erbf}

An \texttt{erbf}\footnote{The ``e'' in \mn{erbf} stands for \emph{ellipsoidal}: each basis function's iso-activation contour forms an axis-aligned ellipsoid when widths differ across dimensions.} places $K$ Gaussian basis functions in the feature space and predicts as a weighted sum of their activations:
\begin{equation}
    f(\mathbf{x}) = \sum_{k=1}^{K} w_k \, \phi_k(\mathbf{x}) + b,
    \qquad
    \phi_k(\mathbf{x}) = \exp\!\Biggl(-\frac{1}{2} \sum_{j=1}^{d} \frac{(x_j - c_{kj})^2}{\sigma_{kj}^2}\Biggr),
    \label{eq:rbf-activation}
\end{equation}
where $\mathbf{c}_k \in \mathbb{R}^d$ is the centre, $\bm{\sigma}_k \in \mathbb{R}^d_{>0}$ is the width vector, $w_k$ is the output weight, and $b$ is a bias.
The key distinction from isotropic RBF networks is that each basis function has a separate width along each feature dimension ($K \times d$ width parameters in total), allowing anisotropic adaptation to local data structure.

Classical RBF training optimised centres and widths simultaneously, typically via gradient descent or metaheuristics, creating a high-dimensional non-convex problem prone to poor local minima.
These training difficulties, together with the limited expressiveness of isotropic widths, contributed to RBF networks falling out of mainstream use.
The pipeline described below reduces the severity of the non-convex optimisation landscape by fixing centres before width optimisation begins, while anisotropic widths adapt to per-feature relevance.

The conceptual insight motivating this decomposition is that centre placement and width adaptation serve distinct purposes: centres determine \emph{where} the model allocates representational capacity, while widths determine \emph{how} that capacity responds to local data structure.
Conflating these in a single optimisation loop (the classical approach) couples two sources of non-convexity: the loss surface contains both the combinatorial difficulty of choosing good locations and the continuous difficulty of shaping the basis functions around them.
Lipschitz-guided centre placement addresses the first problem by exploiting target variation directly: regions where the function changes rapidly need finer coverage, so sampling probability proportional to local Lipschitz estimates concentrates centres where they matter most, without requiring gradient-based search over discrete placements.
With centres fixed, width optimisation becomes a continuous problem in $\mathbb{R}^{K \times d}$ that L-BFGS-B \citep{byrd1995limited} can solve efficiently via analytical gradients.
The decoupling also mitigates a plausible failure mode of joint optimisation: the optimiser may drive widths toward zero around poorly placed centres to fit local noise, producing sharp peaks that overfit.
Fixing centres first constrains width optimisation to refine a geometrically sensible initial configuration rather than compensate for bad placement.

Training proceeds in three decoupled stages.

\subsubsection{Stage 1: Centre placement}
Two strategies are available.
The default, Lipschitz-guided placement, is supervised: we estimate the local Lipschitz constant at each training point as
\begin{equation}
    L_i = \max_{j \in \mathrm{kNN}(i)} \frac{|y_i - y_j|}{\|\mathbf{x}_i - \mathbf{x}_j\| + \epsilon},
\end{equation}
using five nearest neighbours, where $\epsilon$ is a small constant for numerical stability; extreme values are clipped at the 99th percentile for robustness.
The $K$ centres are then sampled without replacement from the training set, with selection probability proportional to $L_i$.
This concentrates centres in high-gradient regions where the target function changes rapidly, allocating modelling capacity where it is most needed.
The $k$-NN-based Lipschitz estimate is inherently noisy, particularly in high-dimensional spaces where neighbour distances become less informative.
In our benchmark, per-fold feature selection limits effective dimensionality to at most 50 features, and the 99th-percentile clipping guards against outlier estimates.
Crucially, the Lipschitz estimate serves only to initialise centre placement; the subsequent width optimisation (Stage~3) compensates for imprecise initial placement.
We verify this robustness empirically (see \cref{sec:lipschitz-sensitivity}).
The alternative, $K$-means clustering, is unsupervised and distributes centres according to feature-space density without accounting for target variation.

\subsubsection{Stage 2: Width initialisation}
Good width initialisation substantially reduces the burden on subsequent optimisation.
Two methods are available:

\emph{Local ridge} (supervised): for each centre $\mathbf{c}_k$, a ridge regression is fitted to its $k$ nearest training points ($k = \min(100,\, \lfloor n/K \rfloor)$, floored at 10), yielding per-feature coefficients $\beta_j$.
The initial width is then
\begin{equation}\label{eq:width-init}
\sigma_{kj} = \tau \sqrt{\frac{\operatorname{Var}(x_j)}{|\beta_j|}},
\end{equation}
where $\operatorname{Var}(x_j)$ is the variance of feature $j$ among the centre's neighbours and $\tau = 1.5\sqrt{d}$ is a dimensional scaling factor (same rationale as for local variance below).
Features with strong local predictive power (large $|\beta_j|$) thus receive narrow widths (high sensitivity), while locally irrelevant features receive wide widths (smoothing).
\emph{Local variance} (unsupervised): widths are set proportional to the local standard deviation of each feature among the centre's neighbours, scaled by $\sqrt{d}$ so that the Gaussian activation remains approximately $e^{-1/2}$ at the typical neighbour distance regardless of dimensionality (the exponent sums $d$ squared terms, so per-dimension widths must grow with $\sqrt{d}$ to compensate).
This purely geometric initialisation ignores target values and is appropriate when features contribute roughly equally.

\subsubsection{Stage 3: Width optimisation}
With centres fixed, widths are optimised by minimising the mean squared error via L-BFGS-B, operating in log-space ($\theta_{kj} = \log \sigma_{kj}$) to enforce positivity.
The width optimisation sub-problem remains non-convex; however, the initialisation from Stage~2 provides a warm start that empirically leads to consistent solutions across random seeds.
We report predictive sensitivity across random seeds in \cref{sec:convergence}.
This log-space parameterisation also reduces the scale sensitivity of the width parameters -- a width change from 0.01 to 0.02 receives the same gradient magnitude as a change from 1.0 to 2.0 -- but does not constitute regularisation in the formal sense.
Analytical gradients are computed in $O(n \cdot K \cdot d)$ per iteration; the activation kernel is JIT-compiled via Numba \citep{lam2015numba} for efficiency.

\subsubsection{Output weights}
Given the activation matrix $\bm{\Phi} \in \mathbb{R}^{n \times K}$, the output weights are obtained by ridge regression,
\begin{equation}\label{eq:ridge-solve}
\mathbf{w} = (\bm{\Phi}^\top \bm{\Phi} + \alpha \mathbf{I}_K)^{-1} \bm{\Phi}^\top \mathbf{y},
\end{equation}
where $\mathbf{I}_K$ is the $K \times K$ identity matrix.

\subsubsection{Automatic scaling of $K$}
Setting the number of centres to its automatic mode applies the empirical heuristic
\begin{equation}\label{eq:auto-k}
K = \operatorname{clip}\bigl(\max(40,\, 2d),\; 20,\; \min(200,\, n/10)\bigr),
\end{equation}
scaling roughly with dimensionality while respecting sample-size and computational constraints.

\subsubsection{Hyperparameters}
The key hyperparameters are the number of centres $K$ (with an automatic mode described above, or a fixed value), the ridge penalty $\alpha$ for the output weights, the centre placement strategy (Lipschitz-guided or $K$-means), and the width initialisation method (local ridge or local variance).
Full implementation details and search spaces are given in \cref{sec:software-details,tab:search-spaces}.

\subsubsection{Training summary}
\Cref{alg:erbf-training} in the appendix summarises the three-stage pipeline.

\subsection{Chebyshev Polynomial Regressor (\mn{chebypoly})}
\label{sec:chebypoly}

The Chebyshev polynomial regressor expands each input feature into a univariate polynomial basis and fits a linear model in the expanded feature space.
We use Chebyshev polynomials of the first kind, $T_n$, defined by the three-term recurrence
\begin{equation}
    T_0(x) = 1, \quad T_1(x) = x, \quad T_{n+1}(x) = 2x\,T_n(x) - T_{n-1}(x),
\end{equation}
which form an orthogonal basis on $[-1,1]$ with respect to the weight $1/\sqrt{1-x^2}$.
Chebyshev bases can improve numerical conditioning relative to monomial bases, although the conditioning of the resulting design matrix depends on the sampled inputs \citep{trefethen2019approximation}.

\subsubsection{Feature expansion}
Given $d$ input features and a maximum degree $c$, we construct a design matrix whose columns are the Chebyshev evaluations
\[
    T_0(x_1),\; T_1(x_1),\; \ldots,\; T_c(x_1),\; T_1(x_2),\; \ldots,\; T_c(x_d),
\]
yielding $1 + d \cdot c$ basis terms (the constant $T_0$ is included only for the first feature to avoid redundant intercepts).
Inputs are first mapped to $[-1,1]$ using a min-max scaler, and values outside the training range are clipped for numerical stability.

\subsubsection{Interaction terms}
Univariate Chebyshev terms capture per-feature nonlinearity but miss bivariate relationships.
When interaction terms are enabled, we augment the basis with pairwise product interactions $x_i \cdot x_j$ between scaled features, adding up to $d(d{-}1)/2$ terms.
Because $T_1(x) = x$, each product coincides with the lowest tensor cross term, $x_i x_j = T_1(x_i)\,T_1(x_j)$, so the default basis is a subset of the tensor-product Chebyshev basis.
That basis is orthogonal with respect to the product Chebyshev measure -- a property of the function system, not of any particular dataset; under the empirical distribution of a given design matrix the columns are not generally orthogonal, especially when features are correlated (\cref{sec:conditioning} reports the resulting empirical conditioning against the monomial basis).
To limit basis size when dimensionality is high ($d > 30$) we restrict interaction pairs to the top half of features ranked by variance.
At the higher interaction complexity setting, each product term is further expanded by appending its Chebyshev evaluation $T_2(x_i \cdot x_j)$, which is not itself a tensor-product basis element; at the default setting, only the raw products are used.

\subsubsection{Regularisation and prediction}
The expanded design matrix is solved by ridge regression.
We tested elastic net regularisation \citep{zou2005regularization} but found that coordinate descent at low $\alpha$ values incurs substantial computational overhead without improving accuracy; ridge proved sufficient.
Predictions are clipped to $\pm 3\sigma$ of the training target to suppress extrapolation artefacts.

\subsubsection{Hyperparameters}
Two hyperparameters govern the core basis expansion: the polynomial degree $c$, which sets the expressiveness of the univariate basis, and the ridge penalty $\alpha$.
Two further parameters control interaction terms: whether to include pairwise product features, and the degree to which those product terms are expanded.
Full implementation details and search spaces are given in \cref{sec:software-details,tab:search-spaces}.

\subsection{Chebyshev Model Tree (\mn{chebytree})}
\label{sec:chebytree}

The Chebyshev model tree is a piecewise-smooth hybrid: a decision tree partitions the feature space into axis-aligned regions, and each leaf fits an independent Chebyshev polynomial regressor.
This combines the tree's capacity for detecting regime boundaries (thresholds, interaction effects, discontinuities) with the smooth local fits provided by Chebyshev polynomials.

\subsubsection{Architecture}
A decision tree regressor is grown with tunable maximum depth and minimum leaf size.
At prediction time, each sample is routed to its leaf by the tree structure, and the leaf-local Chebyshev polynomial generates the prediction.
Leaves with insufficient samples fall back to the tree's own constant prediction to avoid overfitting degenerate local fits.
Each leaf model is fitted independently using ridge regression on the Chebyshev-expanded features of the leaf's training subset.

\subsubsection{Design rationale}
Because each leaf contains only a subset of the training data, leaf polynomials should generally use a lower degree than a global Chebyshev fit to avoid overfitting.
Interaction terms are omitted from leaf models both to reduce overfitting risk and to keep the per-leaf basis size tractable.
The minimum leaf size can be expressed as a fraction of the training set, ensuring leaves remain large enough for stable polynomial fits.

\subsubsection{Hyperparameters}
The model exposes four hyperparameters: the per-leaf polynomial degree, the tree depth, the minimum leaf fraction, and the ridge penalty $\alpha$.
At the extremes of this space, a shallow tree with low-degree polynomials yields a simple piecewise-linear model, while a deep tree with high-degree polynomials produces many nonlinear local surfaces.
Full implementation details and search spaces are given in \cref{sec:software-details,tab:search-spaces}.

\subsection{Explainable Boosting Machine (\mn{ebm})}
\label{sec:ebm}

We included the InterpretML Explainable Boosting Machine (\mn{ebm}) as a bin-based additive comparator \citep{nori2019interpretml}. It assigns a constant contribution within each feature bin and sums these contributions. For computational tractability under nested cross-validation, we disabled interactions (\texttt{interactions}$=0$), which may understate EBM's performance with interactions enabled. We tuned the number of bins, learning rate and minimum samples per leaf (\cref{tab:search-spaces}).

 \subsection{Datasets}
\label{sec:datasets}

We assembled 55 regression datasets from seven sources: OpenML \citep{vanschoren2014openml}, UCI \citep{dua2019uci}, PMLB \citep{romano2021pmlb}, MoleculeNet \citep{wu2018moleculenet}, Hugging Face Datasets, scikit-learn generators, and custom synthetic functions.
Dataset sizes range from $n\!=\!442$ to $581{,}835$ and dimensionality from $d\!=\!2$ to $1{,}024$; Table~S1 in the \href{https://github.com/gerberl/poly-erbf-benchmark/blob/master/paper/supplementary-material.pdf}{Supplementary Material} provides the full list with metadata.

\subsubsection{Stratification}
We grouped the 55 datasets into four application domains to explore whether relative model performance varies by problem type. The groups reflect domain judgement about where smooth responses or thresholds might arise, and individual datasets may differ from those expectations.

\emph{S1: Engineering/Simulation} (13 datasets) includes systems governed by physical laws or smooth equations, such as \dn{airfoil\_noise} and \dn{power\_plant}.

\emph{S2: Behavioural/Social} (10 datasets) includes preferences and demand, such as \dn{wine\_quality} and \dn{Bike\_Sharing\_Demand}. Nine of these datasets have non-continuous targets.

\emph{S3: Physics/Chemistry/Life Sciences} (16 datasets) includes natural-science measurements, such as \dn{superconduct} and \dn{esol}.

\emph{S4: Economics/Pricing} (16 datasets) includes pricing, insurance and real estate, where rules and thresholds may affect the response, as in \dn{house\_sales} and \dn{medical\_charges}.

 \subsection{Benchmark Design and Evaluation Protocol}
\label{sec:benchmark-design}

We applied the evaluation protocol detailed below for nine regression models on the 55 datasets described in \cref{sec:datasets}.
The models span six groups: two smooth-basis models (\mn{chebypoly}, \mn{erbf}); a smooth-tree hybrid (\mn{chebytree}); an additive, bin-based model (\mn{ebm}); two tree ensembles (\mn{rf}, \mn{xgb}); a pre-trained tabular transformer (\mn{tabpfn}); and two baselines at opposite ends of the bias-variance spectrum, linear ridge regression (\mn{ridge}) and decision tree (\mn{dt}).

\subsubsection{Dataset-level preprocessing}
Before splitting, we prepared each dataset using the following steps, without using target values to make preprocessing decisions.
\begin{enumerate}[nosep]
  \item \emph{Featurisation} (dataset-specific). Three molecular datasets distributed as SMILES strings are converted to 16 standard RDKit \citep{landrum2016rdkit} physicochemical descriptors.
  \item \emph{Quality filtering} (all datasets, unconditional). Features with more than 50\% missing values are dropped, as are quasi-constant features (mode frequency $>$95\%).
  \item \emph{Sample reduction.} Datasets exceeding $n\!=\!50{,}000$ are subsampled without replacement, keeping tuning tractable and satisfying the input constraints of \mn{tabpfn}.
\end{enumerate}

\subsubsection{Nested cross-validation}
We used nested cross-validation (CV) to separate hyperparameter tuning from evaluation on held-out data.
\begin{itemize}[nosep]
  \item \emph{Outer loop:} 5-fold CV provides held-out test folds for final evaluation.
  \item \emph{Inner loop:} within each outer training set, hyperparameter search takes place with an adaptive number of inner folds (3-fold CV when $n\!\geq\!1{,}000$, 5-fold otherwise). The selected configuration is then fitted using the outer training data and evaluated on the held-out outer test fold.
\end{itemize}
For models that support early termination (here, XGBoost), 15\% of the outer training set is held out as a validation set for early stopping.

\subsubsection{Fold-level preprocessing}
During tuning, we fit feature selection, imputation, target encoding and any external scaling on each inner training split and apply them to its validation split. After tuning, we fit these steps again on the outer training fold and apply them to the outer test fold.
\begin{itemize}[nosep]
  \item \emph{Feature prefiltering.} For datasets with $d \geq 25$, we keep numeric features with absolute Spearman correlation of at least 0.05. Among the remaining numeric features in the bottom 30\% by absolute Spearman correlation, we retain the top 30\% by mutual information (MI). Categorical features bypass this step.
  \item \emph{Dimensionality reduction.} If $d > 50$ after prefiltering, we retain the 50 features with the highest mutual information.
  \item \emph{Imputation.} Median for numeric features, mode for categorical features.
  \item \emph{Categorical encoding.} Categorical features are converted to numeric via target encoding (\texttt{TargetEncoder} with \texttt{smooth='auto'} via \texttt{sklearn}).
  \item \emph{Standardisation} (model-dependent). \mn{ridge} and \mn{tabpfn} receive standardised features; tree-based models do not; Chebyshev and \texttt{erbf} models apply internal scaling.
\end{itemize}
Selected features may therefore vary across training splits for datasets where feature selection is applied.

\subsubsection{Hyperparameter tuning}
Hyperparameters are optimised with Optuna \citep{akiba2019optuna} using the TPE sampler (\texttt{multivariate=True}) and a \texttt{MedianPruner} for early termination of unpromising trials ($n_{\mathrm{startup}} = 3$, $n_{\mathrm{warmup}} = 1$).
Trial budgets are roughly proportional to search-space dimensionality: \texttt{ridge}~20, \texttt{dt}~25, \texttt{rf}~25, \texttt{xgb}~50, \texttt{erbf}~30, \texttt{chebypoly}~30, \texttt{chebytree}~30, \texttt{ebm}~15; \texttt{tabpfn} uses no tuning (zero-shot).
Search spaces use log-uniform distributions for regularisation parameters, integer ranges for structural parameters (e.g., tree depth, number of centres), and categorical choices for algorithmic variants (e.g., centre placement, width initialisation).
Full per-model search-space definitions are provided in \cref{sec:search-spaces}.
The Supplementary Material summarises the hyperparameter configurations selected across the 55 datasets.

\subsubsection{Evaluation metrics}
We use adjusted $R^2$ ($\bar{R}^2$) as the primary accuracy metric (\cref{sec:results-accuracy}). For each outer fold, we calculate $\bar{R}^2 = 1 - (1 - R^2_{\mathrm{test}})(n_{\mathrm{train}} - 1)/(n_{\mathrm{train}} - d - 1)$, where $n_{\mathrm{train}}$ is the size of the outer training fold and $d$ is the number of input features after preprocessing.
The train--test gap defined in \cref{sec:intro} uses unadjusted $R^2$ on the training and test sets (\cref{sec:results-gap}).
For all models, predictions are clipped to $[y_{\min} - 3\sigma,\; y_{\max} + 3\sigma]$ (computed from the training fold targets) to prevent catastrophic extrapolation from inflating error metrics.
Each metric is aggregated as both mean $\pm$ standard deviation and median across the five outer folds.

\subsubsection{Ranking and statistical testing}
We rank models by adjusted $R^2$ on each dataset and summarise their performance using mean rank and counts of first and second places.\footnote{Ranks and table highlighting use unrounded scores. In descriptive summaries, a missing score receives the worst rank.}
We assess rank differences using the Friedman test followed by Nemenyi pairwise comparisons at $\alpha = 0.05$ \citep{demsar2006statistical}, using datasets with scores for every model in the comparison. The Nemenyi procedure accounts for multiple comparisons.

\subsubsection{Computational setup}
Model--dataset experiments are parallelised across CPU cores via \texttt{joblib}, with the five outer folds within each experiment run sequentially.
To prevent thread thrashing, BLAS thread counts are pinned to~1 (OMP, MKL, and OpenBLAS environment variables) and model-internal parallelism is likewise disabled (\mn{rf} and \mn{xgb}: \texttt{n\_jobs=1}).
Wall-clock times for tuning/setup, training, and prediction are recorded per fold and summed across the five outer folds. Prediction cost is normalised by the total number of training and test rows predicted across those folds and reported per 1\,000 rows.
 \section{Results}
\label{sec:results}

This section reports results on accuracy and train--test gap, together with computational cost, for the eight CPU-viable models.
\mn{tabpfn}, which occupies a distinct category as a pre-trained GPU-dependent transformer, is discussed separately in \cref{sec:results-tabpfn}.
Statistical comparisons use the Friedman test with Nemenyi post-hoc analysis described in \cref{sec:benchmark-design}.

\subsection{Predictive Accuracy}
\label{sec:results-accuracy}

\begin{figure}[htbp]
  \centering
  \begin{subfigure}[t]{0.48\textwidth}
    \centering
    \includegraphics[width=\textwidth]{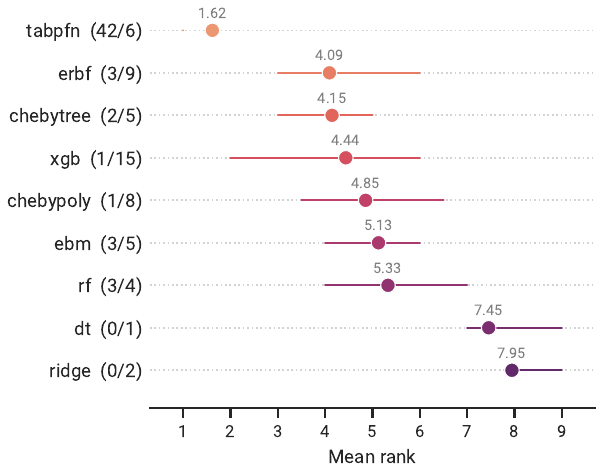}
    \caption{All models}
    \label{fig:rank-all}
  \end{subfigure}
  \hfill
  \begin{subfigure}[t]{0.48\textwidth}
    \centering
    \includegraphics[width=\textwidth]{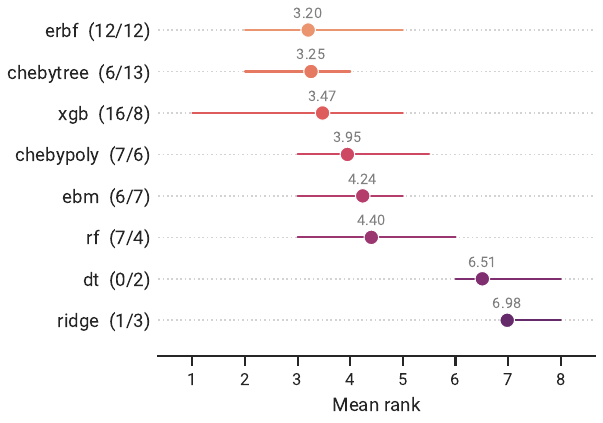}
    \caption{CPU-viable models}
    \label{fig:rank-cpu}
  \end{subfigure}
  \caption{Mean rank for predictive accuracy ($\bar{R}^2$), with counts of first and second places in parentheses. Bars show the interquartile range of per-dataset ranks. The Nemenyi critical difference (\cref{sec:benchmark-design}) is 1.635 for nine models on 54 datasets and 1.416 for eight CPU-viable models on 55 datasets. Full numerical results are given in \cref{tab:accuracy-cpu,tab:accuracy-all} (appendix).}
  \label{fig:rank-combined}
\end{figure}

\Cref{fig:rank-all} shows the mean accuracy ranks for all nine models, including \mn{tabpfn}, which we discuss in \cref{sec:results-tabpfn}.
\Cref{fig:rank-cpu} shows the comparison restricted to the eight CPU-viable models across all 55 datasets.
The top six models -- \mn{erbf}, \mn{chebytree}, \mn{xgb}, \mn{chebypoly}, \mn{ebm}, and \mn{rf} -- are statistically tied (all pairwise rank differences within the Nemenyi critical difference, CD $= 1.416$), with mean ranks between 3.20 and 4.40.
We refer to these six as the \emph{competitive} models hereafter; \mn{dt} and \mn{ridge} form a separate lower group.
The Nemenyi post-hoc test identifies two non-overlapping non-significant groups: \{\mn{erbf}, \mn{chebytree}, \mn{xgb}, \mn{chebypoly}, \mn{ebm}, \mn{rf}\} and \{\mn{dt}, \mn{ridge}\}; every competitive model is significantly better than both \mn{dt} and \mn{ridge}.
\mn{erbf} achieves the best mean rank (3.20) with 12 $R^2$ wins, followed by \mn{chebytree} (3.25; 6 wins) and \mn{xgb} (3.47; 16 wins).

\begin{figure}[htbp]
  \centering
  \includegraphics[width=0.75\textwidth]{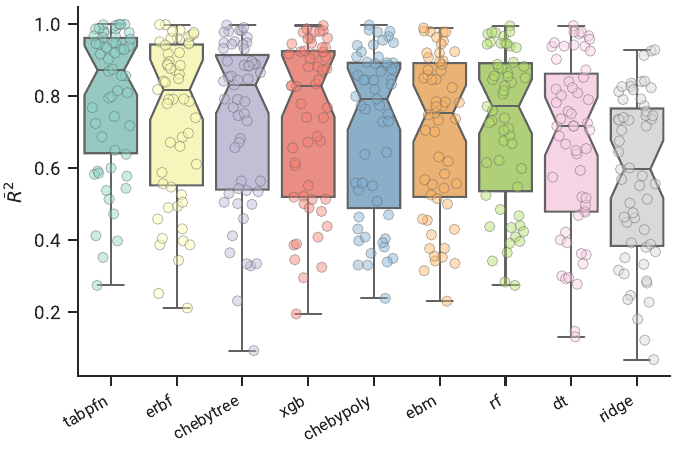}
  \caption{Distribution of $\bar{R}^2$ across 55 datasets per model (ordered by mean rank).  Notched boxes show median and IQR; dots are individual datasets.  The competitive cluster has largely overlapping distributions, with the main separation occurring in the lower tail.}
  \label{fig:boxstrip-r2adj}
\end{figure}

\Cref{fig:boxstrip-r2adj} shows the full distribution of $\bar{R}^2$ across datasets for each model; the six competitive CPU-viable models have largely overlapping distributions.

\subsubsection{Stratum-level analysis}
\label{sec:results-strata}
\Cref{tab:stratum} breaks down mean rank and $\bar{R}^2$ by domain stratum, examining whether model performance varies with the expected characteristics of the data-generating process (see \cref{sec:datasets} for strata definitions).

\begin{table}[htbp]
\centering
\caption{Per-stratum mean rank and $\bar{R}^2$ (all nine models).  Model order as in \Cref{fig:rank-all}.  Within each stratum, the best CPU-viable model per column is in \textbf{bold} and the second \underline{underlined}; rank and $\bar{R}^2$ are emphasised independently.  TabPFN is shown for reference but excluded from the emphasis.}
\label{tab:stratum}
\small
\begin{tabular}{@{}l rr rr rr rr@{}}
\toprule
& \multicolumn{2}{c}{\textbf{S1} (13)} & \multicolumn{2}{c}{\textbf{S2} (10)} & \multicolumn{2}{c}{\textbf{S3} (16)} & \multicolumn{2}{c}{\textbf{S4} (16)} \\
\cmidrule(lr){2-3} \cmidrule(lr){4-5} \cmidrule(lr){6-7} \cmidrule(lr){8-9}
\textbf{Model} & Rank & $\bar{R}^2$ & Rank & $\bar{R}^2$ & Rank & $\bar{R}^2$ & Rank & $\bar{R}^2$ \\
\midrule
\mn{tabpfn} & 1.38 & 0.943 & 3.10 & 0.620 & 1.25 & 0.723 & 1.25 & 0.852 \\
\mn{erbf} & \textbf{3.31} & \textbf{0.917} & 5.40 & 0.546 & \textbf{3.81} & \textbf{0.658} & 4.19 & 0.809 \\
\mn{chebytree} & \underline{3.92} & \underline{0.898} & \textbf{3.80} & \textbf{0.557} & 4.69 & 0.631 & \underline{4.00} & \underline{0.813} \\
\mn{xgb} & \underline{3.92} & 0.875 & 5.30 & 0.551 & 4.94 & \underline{0.652} & \textbf{3.81} & \textbf{0.821} \\
\mn{chebypoly} & 4.54 & 0.827 & \underline{4.30} & 0.550 & \underline{4.50} & 0.630 & 5.81 & 0.788 \\
\mn{ebm} & 6.23 & 0.796 & \underline{4.30} & 0.528 & 4.94 & 0.626 & 4.94 & 0.797 \\
\mn{rf} & 6.08 & 0.870 & 4.90 & \underline{0.552} & 5.06 & 0.636 & 5.25 & 0.793 \\
\mn{dt} & 7.08 & 0.820 & 7.30 & 0.498 & 8.31 & 0.556 & 7.00 & 0.764 \\
\mn{ridge} & 8.54 & 0.674 & 6.60 & 0.398 & 7.50 & 0.545 & 8.75 & 0.636 \\
\bottomrule
\end{tabular}
\end{table}
 
The per-stratum results provide an exploratory description of performance across these domain groupings.

Among CPU-viable models, the pattern is consistent with those expectations.
\mn{erbf} ranks best in S1 and S3 -- the strata where physical or chemical processes suggest smoother target functions -- while \mn{xgb} leads in S4, where threshold-based pricing and policy rules could favour split-based models.
S2 (behavioural/social data) shows lower accuracy across the board ($\bar{R}^2 \approx 0.40$--$0.56$).
The competitive models are tightly bunched in mean $\bar{R}^2$ here (the top five within about 0.01, from 0.546 to 0.557, with \mn{ebm} a little lower at 0.528); \mn{chebytree} leads on mean rank (3.10), ahead of \mn{chebypoly} (3.60) and \mn{ebm} (3.80).

The rank differences, however, are modest (typically less than one rank position between the top CPU-viable models), and the number of datasets per stratum (10--16) limits statistical power.
The stratification should be read as a directional guide rather than a sharp prescription: it suggests which function class (smooth or piecewise-constant) may be more likely to suit a given domain, without guaranteeing that a particular model will win.
Grouping S1 and S3 into a ``smooth'' block (29 datasets) and S2 and S4 into a ``threshold'' block (26 datasets), \mn{erbf} ranks first among CPU-viable models in the smooth block, while \mn{chebytree} leads the threshold block with \mn{xgb} close behind.
The hybrid \mn{chebytree} is competitive across all four strata (rank 3.00--3.75), a pattern consistent with its capacity to combine tree-based regime detection with smooth local fits.

\subsubsection{Effect of target continuity}
We classify a target as non-continuous when it is an intrinsically discrete quantity -- an ordinal rating or satisfaction score, an integer count, or a quantised low-resolution measurement -- rather than a real-valued measurement.
Thirteen of the 55~datasets meet this criterion: six ordinal targets (4--17 levels), four integer counts (10--869 values), and three quantised measurements (35--61 distinct values).
All thirteen take distinct values in at most 5\% of their samples, corroborating their low effective resolution; because the criterion is intrinsic rather than purely cardinality-based, a large dataset whose continuous target merely repeats at coarse resolution (\emph{nyc-taxi}, 0.3\% distinct values from over 580{,}000 samples) is treated as continuous.
\Cref{tab:target-type} compares mean ranks on continuous versus non-continuous targets.

\begin{table}[htbp]
\centering
\caption{Mean rank by target type (CPU-viable models, excluding \mn{tabpfn}).
Non-continuous targets are intrinsically discrete (ordinal ratings, integer counts, or quantised low-resolution measurements; see text for the criterion).
Best in \textbf{bold} per column among competitive models, second \underline{underlined}.}
\label{tab:target-type}
\small
\begin{tabular}{@{}l rr r@{}}
\toprule
\textbf{Model} & \textbf{Continuous (42)} & \textbf{Non-continuous (13)} & \textbf{All (55)} \\
\midrule
\mn{erbf}      & \textbf{2.88} & 4.23 & \textbf{3.20} \\
\mn{chebytree} & 3.40 & \textbf{2.77} & \underline{3.25} \\
\mn{xgb}       & \underline{3.26} & 4.15 & 3.47 \\
\mn{chebypoly} & 4.21 & \underline{3.08} & 3.95 \\
\mn{ebm}       & 4.33 & 3.92 & 4.24 \\
\mn{rf}        & 4.31 & 4.69 & 4.40 \\
\mn{dt}        & 6.43 & 6.77 & 6.51 \\
\mn{ridge}     & 7.17 & 6.38 & 6.98 \\
\bottomrule
\end{tabular}
\end{table}

\mn{erbf} shows the largest rank degradation on non-continuous targets: from 2.88 on continuous data to 4.23 (shift of $+1.35$), winning on none of the non-continuous datasets.
In contrast, \mn{chebypoly} and \mn{chebytree} both \emph{improve} on non-continuous targets (shifts of $-1.13$ and $-0.63$ respectively), with \mn{chebytree} ranking best (2.77).

A plausible architectural explanation is as follows.
\mn{erbf}'s localised Gaussian-like basis functions are a natural match for smooth targets but a poor one for step-like structure, which would require many narrow, overlapping bumps.
\mn{chebypoly}'s global polynomial basis can approximate piecewise structure through oscillation across the domain, reducing its sensitivity to discrete boundaries.
\mn{chebytree} is the most natural fit: its tree routing can partition the feature space into regions of roughly homogeneous target behaviour, while the smooth leaf polynomials handle variation within each partition.
We note, however, that these are post-hoc rationalisations of the rank patterns; the benchmark does not isolate the mechanism.
On the 42~continuous-target datasets, \mn{erbf}'s mean rank improves to 2.88 (from 3.20 overall) and leads the next-best model (\mn{xgb} at 3.26), with all 12 of its wins falling in this subset.

\subsection{Train--test gap}
\label{sec:results-gap}

The train--test gap (training $R^2$ minus test $R^2$) quantifies how much of the training-set fit fails to transfer to held-out data. We examine its relationship with prediction stability under retraining in \cref{sec:results-stability}.
Stability theory provides a motivating link: models whose predictions change little when a single training example is replaced tend to exhibit tighter generalisation bounds \citep{bousquet2002stability}.

Gap magnitudes are not, however, directly comparable across families: different model families control capacity through fundamentally different mechanisms -- early stopping and shrinkage in \mn{xgb}, ridge penalisation in \mn{chebypoly}, basis-function count and width regularisation in \mn{erbf}, pruning depth in \mn{dt}.
The matched-accuracy analysis below restricts comparisons to model pairs with similar held-out performance on each dataset, allowing us to examine differences in training-set fit at comparable test accuracy.
We do not claim that gap constitutes a formal stability measure; rather, it provides a practical diagnostic that complements accuracy.

\Cref{fig:rank-gap} summarises the results (full numerical results in \cref{tab:gap} in the appendix).
\mn{ridge} has the smallest gap (median 0.004), alongside lower predictive accuracy than the six leading CPU models (\cref{sec:results-accuracy}). Its results illustrate why the gap should be considered together with accuracy when comparing models.

Smooth-basis models and the ChebyTree hybrid exhibit substantially tighter train--test gaps than tree ensembles.
Among competitive models, \mn{chebypoly} achieves the best gap rank (3.55), followed by \mn{erbf} (4.27), \mn{ebm} (4.62), and \mn{chebytree} (4.85); \mn{xgb} has the largest gap among competitive models (mean rank 8.31).
These are the nine-model rankings shown in \cref{tab:gap} and \cref{fig:rank-gap} (full numerical results in the appendix).
Among the six competitive CPU models, \mn{chebypoly} achieves 19 gap wins, followed by \mn{chebytree} (10) and \mn{ebm} and \mn{erbf} (9 each).
A tighter gap is consistent with lower training-sample sensitivity, which \cref{sec:results-stability} measures directly under bootstrap retraining.
\mn{ebm}'s train--test gaps are closer to those of the smooth-basis models than to those of the tree ensembles, although its feature contributions are piecewise constant.
At matched accuracy (within 0.02 in $\bar{R}^2$), the smooth-basis models, \mn{chebypoly} and \mn{erbf}, have smaller train--test gaps in 88 of 101 comparisons with \mn{rf} and \mn{xgb} (87\%). The corresponding proportions are 79\% for the \mn{chebytree} hybrid (45/57) and 86\% for \mn{ebm} (37/43), indicating that smaller gaps are not specific to smooth prediction surfaces.
These matched-accuracy counts are descriptive summaries rather than formal hypothesis tests; they complement the Friedman/Nemenyi analysis above by illustrating the direction and consistency of gap differences when accuracy is controlled for.
The pattern is stable across accuracy-matching thresholds spanning an order of magnitude (\cref{tab:threshold-sensitivity} in \cref{sec:appendix-threshold}).

The few gap reversals are concentrated in S4 economic datasets (e.g., \emph{Brazilian\_houses}), where pricing tiers and policy thresholds may create genuinely piecewise-constant structure.
A tree with a small number of well-placed splits can capture such structure at low effective complexity, yielding similar training and test performance; smooth models approximating the same discontinuities may require more basis functions, increasing effective complexity and with it the gap.

\begin{figure}[htbp]
  \centering
  \includegraphics[width=0.55\textwidth]{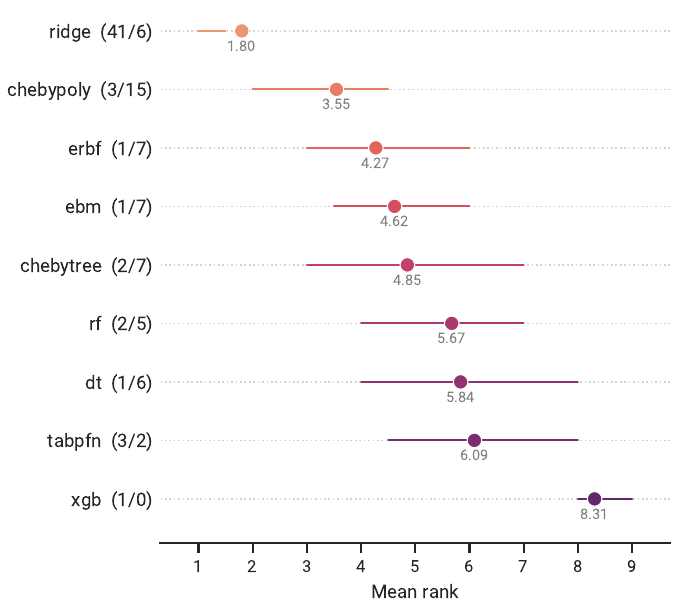}
  \caption{Mean rank for train--test gap ($R^2_{\mathrm{train}} - R^2_{\mathrm{test}}$); parentheses show rank-1 wins\,/\,rank-2 places; bars show the interquartile range.  Lower rank = smaller gap.  Smooth models (\mn{chebypoly}, \mn{erbf}), \mn{ebm}, and the hybrid \mn{chebytree} cluster at the top; \mn{xgb} ranks last among competitive models.  \mn{ridge} has the smallest gap, alongside lower predictive accuracy than the six leading CPU models.}
  \label{fig:rank-gap}
\end{figure}

\subsubsection{Accuracy--gap trade-off}
\begin{figure}[htbp]
  \centering
  \begin{subfigure}[t]{0.48\textwidth}
    \centering
    \includegraphics[width=\textwidth]{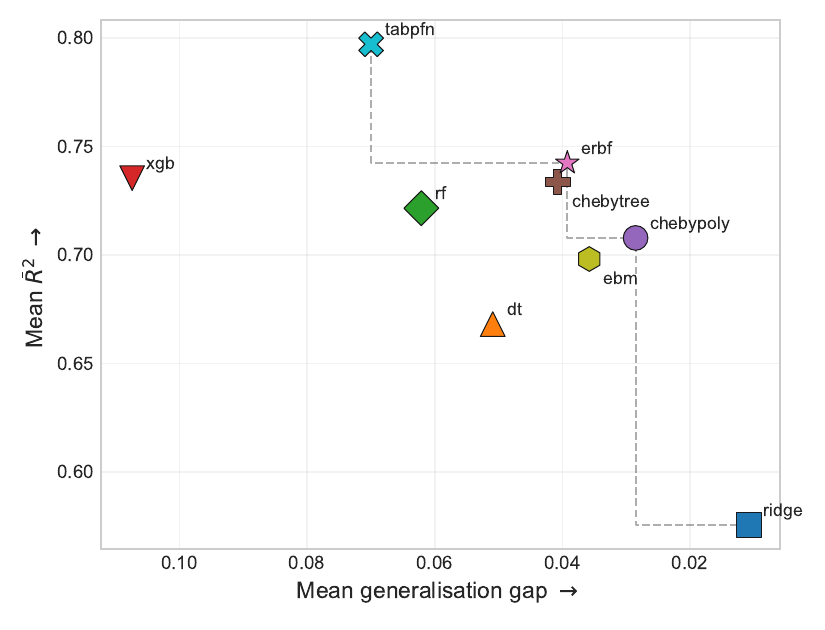}
    \caption{Accuracy vs.\ train--test gap}
    \label{fig:pareto-accuracy-gap}
  \end{subfigure}
  \hfill
  \begin{subfigure}[t]{0.48\textwidth}
    \centering
    \includegraphics[width=\textwidth]{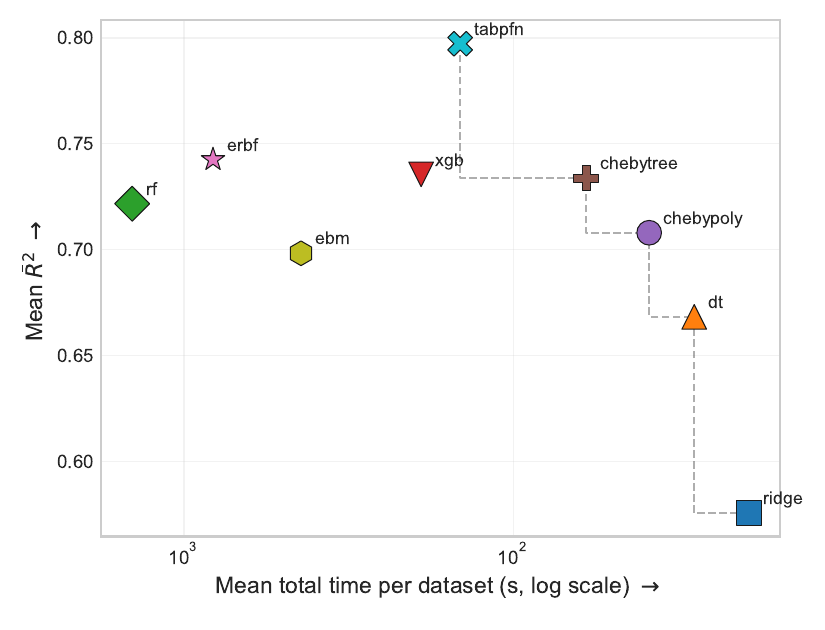}
    \caption{Accuracy vs.\ computational cost}
    \label{fig:pareto-accuracy-time}
  \end{subfigure}
  \caption{Pareto trade-off diagrams.  Each point represents one model at its mean $\bar{R}^2$ plotted against (a)~mean train--test gap and (b)~mean total time per dataset including tuning (log scale).  Step lines trace the Pareto front; arrows on axes indicate the direction of improvement.}
  \label{fig:pareto-combined}
\end{figure}

\Cref{fig:pareto-accuracy-gap} plots each model in ($\bar{R}^2$, gap) space, treating a lower gap descriptively -- as less excess training fit -- as preferable for a given level of accuracy (\cref{fig:pareto-accuracy-time} shows the corresponding accuracy-cost trade-off).
Among CPU-viable models, the Pareto front runs from \mn{ridge} (lowest gap, lowest accuracy) through \mn{chebypoly} and \mn{erbf}; the smooth-basis models occupy the best trade-off positions, achieving competitive accuracy with substantially lower train--test gaps than \mn{xgb} or \mn{rf}.

\subsection{Prediction stability under retraining}
\label{sec:results-stability}

To examine the relationship between the train--test gap and sensitivity to training-sample composition, we refitted each model on 100 bootstrap samples from the first outer training fold and predicted the same held-out cases, keeping feature selection, external preprocessing, tuned hyperparameters and model random seeds fixed.
This measures sensitivity to resampling under the selected configuration, rather than variability from repeating the full selection and tuning procedure.
The analysis covers the eight CPU-viable models on all 55~datasets and \mn{tabpfn} on the 44~datasets within its operating regime.

We summarise stability by a \emph{normalised prediction SD} (SD/span): the median, across test instances, of the standard deviation of their 100 bootstrap predictions, divided by the span between the 1st and 99th percentiles of the training targets.
\Cref{tab:stability} reports the median gap and SD/span per model across the 55~datasets.

\begin{table}[htbp]
\centering
\caption{Median train--test gap and normalised prediction SD (SD/span: the SD of bootstrap predictions relative to the 1st--99th-percentile target span) per model, across the 55 datasets.  Models ordered by SD/span; lower values indicate a smaller train--test gap and more stable predictions.  \mn{tabpfn} is reported separately (44-dataset regime, \cref{sec:results-tabpfn}).}
\label{tab:stability}
\small
\begin{tabular}{@{}lrr@{}}
\toprule
\textbf{Model} & \textbf{Gap (median)} & \textbf{SD/span} \\
\midrule
\mn{ridge}     & 0.004 & 0.007 \\
\mn{chebypoly} & 0.014 & 0.011 \\
\mn{rf}        & 0.024 & 0.013 \\
\mn{chebytree} & 0.026 & 0.018 \\
\mn{ebm}       & 0.020 & 0.023 \\
\mn{erbf}      & 0.018 & 0.025 \\
\mn{xgb}       & 0.079 & 0.030 \\
\mn{dt}        & 0.032 & 0.034 \\
\bottomrule
\end{tabular}
\end{table}

Gap and SD/span are positively associated across the benchmark (Spearman $\rho = 0.78$ over all model--dataset pairs), with a positive association within 95\% of the 55 datasets.
This supports using the gap as a low-cost proxy for retraining stability, although the two measures can give different comparisons.
For example, \mn{dt} and \mn{rf} have median gaps of $0.032$ and $0.024$, yet \mn{rf}'s SD/span is less than half that of \mn{dt} ($0.013$ versus $0.034$), consistent with the variance reduction associated with averaging trees fitted to bootstrap resamples \citep{breiman1996bagging, breiman1996heuristics, linjeon2006adaptive, curth2024forests}.

Restricting to model pairs matched within $0.02~\bar{R}^2$ on each dataset, as in \cref{sec:results-gap}, the smooth-basis models have more stable predictions in 57 of 101 comparisons with the tree ensembles (56\%), compared with smaller gaps in 87\%.
The corresponding stability proportions are 65\% for the \mn{chebytree} hybrid (37/57) and 51\% for \mn{ebm} (22/43).
\mn{chebypoly} and \mn{rf} are among the most stable models, despite belonging to different families; normalisation checks and detailed stability rankings are given in \cref{sec:appendix-stability}.

\subsection{Computational Cost}
\label{sec:results-cost}

Among the competitive CPU-viable models, \mn{chebypoly} and \mn{chebytree} are the cheapest to tune (mean five-fold totals of 38\,s and 59\,s, respectively), since their fitting reduces to ridge regression in the expanded basis.
\mn{xgb} (189\,s) is moderate; \mn{ebm} (432\,s), \mn{erbf} (800\,s), and \mn{rf} (1\,395\,s) are substantially more expensive, driven by cyclic boosting iterations, L-BFGS-B width optimisation, and full forest fitting per Optuna trial, respectively.
Once tuned, \mn{ebm} has the lowest mean batch prediction cost among competitive models (0.46\,ms per 1\,000 rows), followed by \mn{erbf} (2.12\,ms), \mn{xgb} (3.35\,ms), and \mn{chebypoly} (3.48\,ms); all CPU models average under 8\,ms per 1\,000 rows except \mn{rf} (51.15\,ms).
Full timing breakdowns are in \cref{tab:timing,tab:timing-examples} in the appendix.

The accuracy-cost Pareto front (\cref{fig:pareto-accuracy-time}) shows \mn{ridge}, \mn{dt}, \mn{chebypoly}, and \mn{chebytree} at the fast end; \mn{ebm} occupies an intermediate position with moderate tuning cost but very fast prediction.
\mn{erbf} and \mn{rf} fall off the front due to their higher tuning cost.

\subsubsection{Scalability check}
The main benchmark uses preprocessed data (at most 50 features and 50\,000 samples) to keep computational cost manageable for \mn{erbf} and \mn{rf} tuning at high dimensionality.
These caps bind on eight of the 55 datasets -- five subsampled to 50\,000 instances and four reduced to 50 features -- so the great majority enter tuning at full size.
To test whether the Chebyshev models remain competitive at full scale, we re-ran the five scalable models (\mn{ridge}, \mn{dt}, \mn{xgb}, \mn{chebypoly}, and \mn{chebytree}) on eight datasets at full scale -- all features and samples, no feature selection or subsampling -- with up to $n = 581{,}835$ samples and $d = 1{,}024$ features, and place the full-scale scores alongside the corresponding main-benchmark (reduced) values in \cref{tab:scalability-check}.

\begin{table}[ht]
\centering
\caption{Scalability check: adjusted $R^2$ ($\bar{R}^2$) for the five scalable models on eight datasets, comparing the main-benchmark \emph{reduced} setting (\textbf{r}; at most 50 features and 50\,000 samples, feature selection refit per fold) with the \emph{full}-scale run (\textbf{f}; all features and samples, no feature selection or subsampling).  $n$ and $d$ are the full-scale sample and feature counts.  Best full-scale value per dataset in \textbf{bold}; datasets ordered by mean full-scale $\bar{R}^2$.}
\label{tab:scalability-check}
\scriptsize
\setlength{\tabcolsep}{3.5pt}
\begin{tabular}{@{}lrr *{5}{cc}@{}}
\toprule
 & & & \multicolumn{2}{c}{\mn{xgb}} & \multicolumn{2}{c}{\mn{chebytree}} & \multicolumn{2}{c}{\mn{chebypoly}} & \multicolumn{2}{c}{\mn{ridge}} & \multicolumn{2}{c}{\mn{dt}} \\
\cmidrule(lr){4-5}\cmidrule(lr){6-7}\cmidrule(lr){8-9}\cmidrule(lr){10-11}\cmidrule(lr){12-13}
\textbf{Dataset} & \textbf{$n$} & \textbf{$d$} & r & f & r & f & r & f & r & f & r & f \\
\midrule
medical\_charges & 163K & 3    & 0.979 & 0.978 & 0.980 & \textbf{0.980} & 0.980 & 0.979 & 0.827 & 0.825 & 0.978 & 0.978 \\
diamonds         &  54K & 6    & 0.945 & \textbf{0.946} & 0.943 & 0.945 & 0.943 & 0.943 & 0.925 & 0.926 & 0.942 & 0.942 \\
superconduct     &  21K & 79   & 0.909 & \textbf{0.910} & 0.890 & 0.900 & 0.819 & 0.837 & 0.675 & 0.733 & 0.819 & 0.825 \\
friedman1\_d100  &  2K  & 100  & 0.904 & \textbf{0.926} & 0.889 & 0.878 & 0.875 & 0.905 & 0.736 & 0.710 & 0.716 & 0.676 \\
geo\_music       &  1K  & 117  & 0.675 & 0.647 & 0.731 & 0.670 & 0.774 & \textbf{0.758} & 0.774 & 0.756 & 0.521 & 0.473 \\
qsar\_tid\_11    &  6K  & 1024 & 0.607 & \textbf{0.665} & 0.535 & 0.649 & 0.398 & 0.523 & 0.317 & 0.498 & 0.393 & 0.420 \\
allstate\_claims & 188K & 130  & 0.480 & 0.525 & 0.500 & \textbf{0.526} & 0.470 & 0.482 & 0.450 & 0.462 & 0.401 & 0.407 \\
nyc\_taxi        & 582K & 9    & 0.513 & \textbf{0.580} & 0.530 & 0.568 & 0.393 & 0.398 & 0.307 & 0.306 & 0.517 & 0.513 \\
\midrule
\multicolumn{3}{@{}l}{\emph{Mean}} & 0.752 & 0.772 & 0.750 & 0.764 & 0.707 & 0.728 & 0.626 & 0.652 & 0.661 & 0.654 \\
\bottomrule
\end{tabular}
\end{table}

Two points emerge, subject to the small sample.
First, both Chebyshev models scale: \mn{chebytree} tracks \mn{xgb} on the mean (full-scale $\bar{R}^2$ 0.764 vs.\ 0.772), and \mn{chebypoly}, though lower at 0.728, stays clear of \mn{ridge} and \mn{dt}, with both models completing the full-scale runs, including datasets with up to 581,835 rows or 1,024 features.
Second, the main-benchmark caps do not tilt the smooth-versus-tree comparison.
On the low-dimensional datasets \emph{medical\_charges} and \emph{diamonds}, accuracy is essentially unchanged between the reduced and full settings, and where full scale does lift scores the gain is shared rather than tree-favouring: on \emph{qsar\_tid\_11}, whose feature count is capped in the main benchmark, the smooth models gain at least as much as \mn{xgb} (\mn{chebypoly} $+0.13$ vs.\ \mn{xgb} $+0.06$ in $\bar{R}^2$).
\mn{xgb} is not disadvantaged by the caps -- it improves at full scale on the genuinely larger problems (\emph{nyc\_taxi} $+0.07$, \emph{allstate\_claims} and \emph{qsar\_tid\_11} $+0.05$--$0.06$) -- but so do the smooth models, and the only change in the highest-scoring model occurs on \emph{nyc\_taxi}, where \mn{xgb} overtakes \mn{chebytree} at full scale.
The limited number of datasets precludes strong conclusions, but the preprocessing caps appear neither to inflate nor to deflate the Chebyshev models' standing relative to the tree ensembles.

\subsection{Comparison with TabPFN}
\label{sec:results-tabpfn}

The preceding subsections focused on the eight CPU-viable models.
\mn{tabpfn} \citep{hollmann2025tabpfn} occupies a distinct position: it is a pre-trained transformer that requires no per-dataset hyperparameter tuning and runs on GPU, differing fundamentally from the fitted models evaluated above.
Our benchmark centres on models that train and predict on commodity hardware without GPU acceleration -- the CPU-based settings common across applied science and engineering -- so \mn{tabpfn}'s reliance on GPU inference places it outside that primary scope.
We therefore collect its results here as a separate reference point rather than interleaving them with the CPU-viable analysis.

\subsubsection{Accuracy}
When included in the nine-model ranking (\cref{fig:rank-all}; full numerical results in \cref{tab:accuracy-all}), \mn{tabpfn} achieves 42 rank-1 placements out of 54 datasets and a mean rank of 1.62, placing it ahead of the competitive CPU-viable group.\footnote{One dataset (\dn{food\_delivery\_time}) is excluded from the nine-model comparison because \mn{tabpfn} triggered a GPU out-of-memory error; its rank on that dataset is set to worst (9).}
Cross-dataset standard deviation of $\bar{R}^2$ is slightly lower for \mn{tabpfn} (0.20) than for the six competitive CPU models (0.22--0.23).
The stratum breakdown (\cref{tab:stratum}) shows \mn{tabpfn} leading in all four strata, with its narrowest margin in S2 (behavioural/social data, rank 3.10 vs.\ 3.80 for the next-best CPU model), where only 9 of 10 datasets are available.

\subsubsection{Train--test gap}
\mn{tabpfn}'s gap (median 0.047, mean rank 6.09; \cref{fig:rank-gap}) falls between \mn{dt} and \mn{xgb}, closer to the tree ensembles than to the smooth models.
\mn{tabpfn} conditions its predictions on the training examples without updating its pre-trained parameters. We report its train--test gap descriptively, without attributing it to a particular mechanism.
The accuracy--gap Pareto front (\cref{fig:pareto-accuracy-gap}) places \mn{tabpfn} at the high-accuracy end with a moderate gap.

\subsubsection{Operational constraints}
\mn{tabpfn} requires no hyperparameter tuning but incurs the highest mean batch prediction cost among all models tested (1\,253\,ms per 1\,000 rows across its 54 datasets, compared with under 8\,ms for all CPU models except \mn{rf} at 51.15\,ms).
Its GPU memory consumption scales as $O(n^2)$ due to attention over training instances, with a practical ceiling of 50\,000 samples; one dataset in our benchmark (\dn{food\_delivery\_time}, $n \approx 45$K) triggered a CUDA out-of-memory error.
These constraints restrict its applicability to moderate-sized problems where GPU inference is available.

Tabular deep learning is an active and rapidly evolving area; a systematic comparison across architectures (e.g., TabPFN, FT-Transformer, GRANDE) would be a valuable but separate study.
Our inclusion of \mn{tabpfn} as a single representative provides a reference point for the accuracy frontier while keeping the benchmark's primary focus on CPU-viable model families.

\subsection{Practical Model Selection}
\label{sec:taxonomy}

The preceding subsections evaluated models along individual axes -- accuracy, train--test gap, and computational cost.
Here we draw these threads together into practical guidance for model selection.
No single model dominates all criteria, and the best choice depends on which trade-offs matter most for a given application; all competitive families should be benchmarked on the target problem.
What follows are empirically grounded priors to inform that process.

Domain knowledge can inform priors about which family to favour.
The stratum analysis (\cref{sec:results-strata}) offers a directional guide: smooth models tend to rank higher on datasets from engineering, simulation, and physical-science domains (S1, S3), while tree ensembles tend to lead on economic and pricing datasets with threshold-driven structure (S4).
These are tendencies, not guarantees (\cref{sec:datasets}).

When models achieve comparable accuracy on a given problem -- which, given the aggregate tie, can often be the case -- other criteria become decisive.
The smooth-basis matched-accuracy gap advantage (87\% of pairwise comparisons, \cref{sec:results-gap}) and the accuracy-gap Pareto front (\cref{fig:pareto-accuracy-gap}) both favour smooth models.
Retraining stability is more equivocal: at matched accuracy, the smooth-basis models are more stable in 56\% of comparisons with tree ensembles (\cref{sec:results-stability}). \mn{chebypoly} and \mn{rf} are among the most stable models, despite belonging to different families.
Depending on the application, a practitioner may also choose to accept a small accuracy trade-off in favour of smooth prediction surfaces (for surrogate optimisation or sensitivity analysis), structural interpretability, or lower computational cost.

On computational cost (\cref{sec:results-cost}), \mn{chebypoly} and \mn{chebytree} are the fastest competitive models, placing them on the accuracy-cost Pareto front (\cref{fig:pareto-accuracy-time}).
\mn{erbf} and \mn{rf} fall off this front due to expensive tuning, though \mn{erbf}'s fast inference (\cref{tab:timing}) means the tuning cost is a one-time investment amortised over subsequent predictions.

Scalability further differentiates the models.
The scalability check (\cref{tab:scalability-check}) showed that \mn{xgb}, \mn{chebytree}, and \mn{chebypoly} all run at full scale without preprocessing, though the Chebyshev basis size grows combinatorially with dimensionality.
\mn{erbf}'s $O(nKd)$ per-iteration cost makes it the most expensive smooth model to tune at high $n$ or $d$.

Smooth models also offer a degree of structural interpretability not shared by tree ensembles.
\mn{chebypoly} yields explicit polynomial coefficients from which partial derivatives and sensitivity indices can be computed; \mn{chebytree} combines tree routing with local polynomial coefficients; \mn{erbf} provides geometric interpretability via localised basis functions whose widths indicate local feature relevance.
\mn{xgb} and \mn{rf} are more commonly interpreted through post-hoc tools (SHAP, partial dependence), though these tools are equally applicable to smooth models.
These are structural properties of the model architectures; a comparative evaluation of interpretability is beyond the scope of this benchmark.

In summary, when train--test gap and interpretability are priorities, \mn{erbf} and \mn{chebypoly} offer the strongest profiles among CPU-viable models.
When stable predictions under retraining matter, \mn{chebypoly} and \mn{rf} performed best among the competitive CPU models in this benchmark.
When tuning budget is limited, \mn{chebytree} and \mn{chebypoly} provide competitive accuracy at the lowest cost.
When scalability to high-dimensional data is the binding constraint, \mn{xgb} and \mn{chebytree} both proved viable at full scale.
Where GPU inference is available, \mn{tabpfn} can extend the accuracy frontier further (\cref{sec:results-tabpfn}).

 \section{Discussion}
\label{sec:discussion}

\subsection{Scope and limitations}
The benchmark is restricted to regression; classification tasks may favour different model families.

Our evaluation uses a fixed nested cross-validation protocol with Optuna-based hyperparameter tuning.
Trial budgets (20--50 per model, proportional to search space dimensionality) were chosen to keep resource usage comparable across models; more trials could shift relative rankings, particularly for \mn{erbf}, whose width optimisation involves a non-convex inner loop that benefits from broader search.

The main preprocessing pipeline applies mutual-information-based feature selection (capping at 50 features) and sample subsampling (capping at 50\,000 instances) to keep nested-CV tuning tractable across all model families; the same thresholds conveniently satisfy \mn{tabpfn}'s input constraints.
Although these caps are applied uniformly, their effect on different model families is not necessarily symmetric.
The scalability check (\cref{tab:scalability-check}) partially addresses this by running scalable models on full-scale data without the sample and feature caps, but a complete comparison at full scale for all models would require substantially more computation.
Relatedly, we applied a uniform preprocessing pipeline across all datasets rather than tailoring feature engineering to each domain; domain-specific feature construction (e.g.\ task-specific molecular fingerprints instead of generic RDKit descriptors) might improve all models.
Results on individual datasets should therefore be interpreted as reflecting the model-pipeline combination, not the model alone.

ERBF's training complexity is $O(nKd)$ per L-BFGS-B iteration, where $n$ is the number of samples, $K$ the number of centres, and $d$ the dimensionality.
In our experiments, ERBF trained comfortably on datasets up to approximately 10\,000 samples and 50 features on a standard multi-core CPU, with mean tuning time of 800\,s compared to 189\,s for XGBoost.
For larger problems, the absence of mini-batch support and the growth of the kernel matrix with $K$ become limiting factors; extending ERBF with stochastic optimisation or inducing-point approximations is a natural direction for future work.

All train-from-scratch models in the benchmark operate in axis-aligned feature coordinates: trees splits act on single features, ERBF uses diagonal (per-dimension) widths, and Chebyshev interaction terms are products along coordinate axes.
Data with strong oblique structure (relationships aligned along linear combinations of features) may affect the model families differently; preprocessing such as PCA rotation could benefit all models.

The train--test gap reflects model capacity, regularisation and optimisation as well as the data. Our matched-accuracy analysis (\cref{sec:results-gap}) compares models at similar held-out performance, but does not control these other influences.

The 55 datasets span four domain strata, but provide limited evidence for conclusions about individual application domains.
The stratum-level analysis (\cref{sec:results-strata}) is exploratory rather than confirmatory.

\subsection{Future work}
Extending the evaluation to classification is the most natural next step.
Formal assessment of prediction-surface regularity, for example via local Lipschitz analysis, could complement the train--test gap analysis presented here.
For ERBF specifically, scaling the $O(nKd)$ per-iteration cost via mini-batch optimisation or sparse activations remains an open direction.
Hybrid architectures that combine tree-based routing with local smooth fits, or target transformations that map discrete outcomes to a continuous latent space, could address ERBF's sensitivity to non-continuous targets.

\subsection{Broader implications}
These results support including smooth-basis models among the candidates for tabular regression.
Across this benchmark, they are competitive with tree ensembles on accuracy while offering different trade-offs in train--test gap, computational cost and prediction-surface smoothness.
The exploratory stratum analysis (\cref{sec:results-strata}) provides some guidance on where they may be useful: \mn{erbf} ranks highest among CPU-viable models in the engineering and natural-science strata, while the \mn{chebytree} hybrid remains competitive across all four strata.
These patterns do not establish domain-specific rules, since the groups are small and their construction already reflects expectations about the underlying relationships.

Accuracy alone does not describe all differences relevant to model choice.
Models with similar held-out scores can differ in both training-set fit and sensitivity to changes in the training sample.
Our bootstrap analysis (\cref{sec:results-stability}) finds a positive association between the train--test gap and prediction stability under retraining, supporting the gap as an inexpensive diagnostic alongside accuracy.
The measures nevertheless give different comparisons for some models, and neither a small gap nor a smooth prediction surface guarantees stable predictions under retraining.

Smooth prediction surfaces may also be useful independently of their relationship with the train--test gap.
In surrogate-based engineering design, differentiable models allow gradient-based optimisation and sensitivity analysis \citep{forrester2009recent,jones1998efficient}.
Where small changes in continuous inputs should produce gradual changes in predictions, smooth-basis models offer an alternative to piecewise-constant predictions.
This consideration applies to \mn{erbf} and \mn{chebypoly}; \mn{chebytree} provides smooth fits within leaves but can retain discontinuities between them.
Together with accuracy and computational cost, these properties provide practical reasons to consider a wider range of regression models.

\section{Conclusion}
\label{sec:conclusion}

We have benchmarked two smooth-basis model families (Chebyshev polynomial regressors and anisotropic RBF networks) and a smooth-tree hybrid (Chebyshev model trees) against tree ensembles, a bin-based additive model, a pre-trained transformer and standard baselines across 55 tabular regression datasets.
We evaluated predictive accuracy, the train--test gap and computational cost, and used bootstrap retraining to examine prediction stability.

The main finding is that smooth-basis models are competitive with tree ensembles on predictive accuracy while exhibiting tighter train--test gaps: the six competitive CPU-viable models are statistically indistinguishable on accuracy, but the smooth-basis models show smaller train--test gaps in the large majority of pairwise comparisons at matched accuracy (\cref{sec:taxonomy}).
At matched test accuracy, the smaller gap reflects lower training-set fit; its positive association with prediction stability under retraining supports reporting it alongside accuracy.
These findings apply primarily to the moderate sample sizes and feature counts represented in the main benchmark ($n \lesssim 50{,}000$, $d \lesssim 50$).
The full-scale check on eight datasets gives similar results overall for the five scalable models evaluated, although \mn{xgb} substantially outperforms \mn{chebypoly} on \emph{nyc\_taxi}, the largest dataset (\cref{tab:scalability-check}).

These results support routinely including smooth-basis models among the candidates for tabular regression.
When models achieve comparable accuracy, prediction stability, smoothness of the prediction surface and computational cost can help guide the choice according to the application.

\subsection{Reproducibility}
All experiments use fixed random seeds (outer CV seed 42, Optuna seed 0) and deterministic nested cross-validation splits.
The model implementations are available on PyPI (\texttt{pip install erbf poly-basis-ml}), and the full benchmark code--including analysis scripts, dataset loaders, and result summaries--is available at \url{https://github.com/gerberl/poly-erbf-benchmark}.
The benchmark was run with Python 3.12, scikit-learn 1.6, XGBoost 3.1, Optuna 4.6, and NumPy 2.3; a complete environment specification (\texttt{environment.yml}) is included in the repository.
Datasets are drawn from OpenML, UCI, PMLB, MoleculeNet, Hugging Face Datasets, scikit-learn generators, and custom synthetic functions (sources and IDs listed in the \href{https://github.com/gerberl/poly-erbf-benchmark/blob/master/paper/supplementary-material.pdf}{Supplementary Material}).
Experiments were conducted on a workstation with an Intel Core i9-9900X CPU (10 cores, 20 threads, 3.50\,GHz), 128\,GB RAM, and an NVIDIA TITAN RTX GPU (24\,GB VRAM), running Linux 6.1.
Hyperparameter search spaces are fully specified in \cref{sec:search-spaces}, and per-dataset results are reported in \cref{sec:per-dataset-results}.

\bibliography{references}

\appendix

\section{ERBF Training Pipeline}
\label{sec:erbf-algorithm}

\begin{algorithm}[H]
\caption{Three-stage ERBF training pipeline.}
\label{alg:erbf-training}
\begin{algorithmic}[1]
\Require Training data $\{(\mathbf{x}_i, y_i)\}_{i=1}^n$, number of centres $K$, ridge penalty $\alpha$, neighbourhood size $k_{\mathrm{nn}}$
\Ensure Centres $\{\mathbf{c}_k\}$, widths $\{\bm{\sigma}_k\}$, weights $\{w_k\}$, bias $b$
\Statex \textbf{Stage 1: Centre placement} \Comment{Hyperparameter: \texttt{center\_init}}
\If{Lipschitz-guided (default)}
    \For{$i = 1, \ldots, n$}
        \State Compute local Lipschitz constant $L_i = \max_{j \in \mathrm{kNN}(i)} \frac{|y_i - y_j|}{\|\mathbf{x}_i - \mathbf{x}_j\| + \epsilon}$
    \EndFor
    \State Clip $L_i$ at the 99th percentile
    \State Sample $K$ centres $\{\mathbf{c}_k\}$ from training points without replacement, with $P(\mathbf{x}_i) \propto L_i$
\ElsIf{$K$-means}
    \State Run $K$-means on training features $\to$ centroids $\{\mathbf{c}_k\}$
\EndIf
\Statex \textbf{Stage 2: Width initialisation} \Comment{Hyperparameter: \texttt{width\_init}}
\For{$k = 1, \ldots, K$}
    \State Find $k_{\mathrm{nn}}$ nearest neighbours of $\mathbf{c}_k$
    \If{Local ridge (default)}
        \State Fit local ridge regression $\to$ coefficients $\beta_j$
        \State Set $\sigma_{kj}^2 \gets \mathrm{Var}_{\mathrm{local}}(x_j) \,/\, |\beta_j|$ for $j = 1, \ldots, d$
    \ElsIf{Local variance}
        \State Set $\sigma_{kj} \gets \mathrm{std}_{\mathrm{local}}(x_j) \cdot \sqrt{d}$ for $j = 1, \ldots, d$
    \EndIf
\EndFor
\Statex \textbf{Stage 3: Width optimisation}
\State $\bm{\theta} \gets \log \bm{\sigma}$ \Comment{Enforce positivity via log-space}
\State $\bm{\theta}^* \gets \text{L-BFGS-B}\bigl(\min_{\bm{\theta}} \tfrac{1}{n}\sum_{i=1}^n (y_i - f(\mathbf{x}_i))^2,\; \bm{\theta}_0 = \bm{\theta},\; \text{maxiter}=30\bigr)$
\State $\bm{\sigma} \gets \exp(\bm{\theta}^*)$
\Statex \textbf{Output weights}
\State Compute activation matrix $\bm{\Phi} \in \mathbb{R}^{n \times K}$ with $\Phi_{ik} = \phi_k(\mathbf{x}_i)$ \Comment{Eq.~\ref{eq:rbf-activation}}
\State $\mathbf{w} \gets (\bm{\Phi}^\top \bm{\Phi} + \alpha \mathbf{I})^{-1} \bm{\Phi}^\top \mathbf{y}$ \Comment{$\mathbf{I} \in \mathbb{R}^{K \times K}$ identity matrix}
\end{algorithmic}
\end{algorithm}

\section{Hyperparameter Search Spaces}
\label{sec:search-spaces}

\Cref{tab:search-spaces} lists the full per-model search spaces used by Optuna during the inner loop of nested cross-validation.
All continuous regularisation parameters are sampled on a log-uniform scale; structural parameters (depths, counts) use uniform integer sampling; algorithmic variants use categorical sampling.
Fixed (non-tuned) settings are shown as greyed rows.

{\scriptsize
\setlength{\tabcolsep}{3pt}
\begin{longtable}{@{}llll@{}}
\caption{Hyperparameter search spaces and fixed settings for each model.  ``Log'' indicates log-uniform sampling; ``Cat'' indicates categorical.  Trial budgets are given in parentheses after each model name.  Greyed rows show fixed (non-tuned) settings.}
\label{tab:search-spaces} \\
\toprule
\textbf{Model} & \textbf{Parameter} & \textbf{Range / choices} & \textbf{Scale} \\
\midrule
\endfirsthead
\toprule
\textbf{Model} & \textbf{Parameter} & \textbf{Range / choices} & \textbf{Scale} \\
\midrule
\endhead
\midrule \multicolumn{4}{r}{\emph{Continued on next page}} \\
\endfoot
\bottomrule
\endlastfoot
\textbf{Ridge} (20 trials)
  & \texttt{alpha} & $[10^{-3},\; 10^{3}]$ & Log \\
\midrule
\textbf{Decision Tree} (25 trials)
  & \texttt{max\_depth} & $[1,\; 20]$ & Int \\
  & \texttt{min\_samples\_leaf} & $[0.005,\; 0.1]$ & Uniform \\
  & \texttt{min\_samples\_split} & $[0.01,\; 0.1]$ & Uniform \\
\midrule
\textbf{Random Forest} (25 trials)
  & \texttt{n\_estimators} & $[50,\; 500]$ & Int \\
  & \texttt{max\_depth} & $[3,\; 20]$ & Int \\
  & \texttt{max\_features} & \{0.3, 0.5, 0.7, sqrt\} & Cat \\
  \rowcolor{black!5}
  & \texttt{min\_samples\_leaf} & 0.005 & Fixed \\
\midrule
\textbf{XGBoost} (50 trials)
  & \texttt{max\_depth} & $[1,\; 9]$ & Int \\
  & \texttt{learning\_rate} & $[0.01,\; 0.3]$ & Log \\
  & \texttt{subsample} & $[0.6,\; 1.0]$ & Uniform \\
  & \texttt{colsample\_bytree} & $[0.6,\; 1.0]$ & Uniform \\
  & \texttt{reg\_lambda} & $[10^{-3},\; 10^{3}]$ & Log \\
  & \texttt{min\_child\_weight} & $[1,\; 100]$ & Log \\
  \rowcolor{black!5}
  & \texttt{n\_estimators} & 2\,000 (early stop, patience 10) & Fixed \\
  \rowcolor{black!5}
  & validation split & 15\% of training set & Fixed \\
\midrule
\textbf{ERBF} (30 trials)
  & \texttt{n\_rbf} & \{auto\} $\cup$ $[10,\; 80]$ & Cat / Int \\
  & \texttt{alpha} & $[10^{-3},\; 10^{3}]$ & Log \\
  & \texttt{center\_init} & \{lipschitz, kmeans\} & Cat \\
  & \texttt{width\_init} & \{local\_ridge, local\_variance\} & Cat \\
  \rowcolor{black!5}
  & \texttt{width\_optim\_iters} & 30 & Fixed \\
  \rowcolor{black!5}
  & \texttt{width\_mode} & full (gradient-based) & Fixed \\
\midrule
\textbf{ChebyPoly} (30 trials)
  & \texttt{complexity} & $[1,\; 14]$ & Int \\
  & \texttt{alpha} & $[10^{-3},\; 10^{3}]$ & Log \\
  & \texttt{include\_interactions} & \{True, False\} & Cat \\
  & \texttt{max\_interaction\_complexity} & \{1, 2\} & Cat$^{\dagger}$ \\
  \rowcolor{black!5}
  & \texttt{solver} / \texttt{clip\_input} & Ridge; True & Fixed \\
  \rowcolor{black!5}
  & \texttt{interaction\_types} & [product] & Fixed \\
\midrule
\textbf{ChebyTree} (30 trials)
  & \texttt{complexity} & $[1,\; 6]$ & Int \\
  & \texttt{alpha} & $[10^{-3},\; 10^{3}]$ & Log \\
  & \texttt{max\_depth} & $[1,\; 12]$ & Int \\
  & \texttt{min\_samples\_leaf} & $[0.01,\; 0.1]$ & Uniform \\
  \rowcolor{black!5}
  & \texttt{solver} & Ridge & Fixed \\
\midrule
\textbf{EBM} (15 trials)
  & \texttt{max\_bins} & \{128, 256\} & Cat \\
  & \texttt{learning\_rate} & $[0.01,\; 0.1]$ & Log \\
  & \texttt{min\_samples\_leaf} & \{2, 4, 10\} & Cat \\
  \rowcolor{black!5}
  & \texttt{interactions} & 0 (pure additive) & Fixed \\
  \rowcolor{black!5}
  & \texttt{max\_leaves} & 3 & Fixed \\
  \rowcolor{black!5}
  & \texttt{outer\_bags} & 4 & Fixed \\
  \rowcolor{black!5}
  & \texttt{max\_rounds} & 5\,000 & Fixed \\
  \rowcolor{black!5}
  & \texttt{early\_stopping\_rounds} & 50 & Fixed \\
\midrule
\textbf{\mn{tabpfn}} (0 trials)
  & \multicolumn{3}{l}{Zero-shot; no hyperparameters tuned.} \\
\end{longtable}

\smallskip
\noindent{\footnotesize
$^{\dagger}$\,Conditional on \texttt{include\_interactions}~=~True.
All stochastic models use \texttt{random\_state}~=~42.
}
}

\section{Implementation and Reproducibility}
\label{sec:software-details}

All three contributed models are implemented in Python, building on NumPy \citep{harris2020numpy}, SciPy \citep{virtanen2020scipy}, scikit-learn \citep{pedregosa2011scikit}, and joblib \citep{joblib} for parallelisation.
The Chebyshev models are provided by the \texttt{poly\_basis\_ml} package and the RBF network by the \texttt{erbf} package; both expose scikit-learn-compatible estimators (\texttt{fit}\slash\texttt{predict} interface) and are available on PyPI (\texttt{pip install erbf poly-basis-ml}) and GitHub.\footnote{\url{https://github.com/gerberl/erbf} and \url{https://github.com/gerberl/poly-basis-ml}}

The tuned hyperparameters for all three contributed models, with their ranges and fixed settings, are listed in \cref{tab:search-spaces}; we note here only the implementation details not captured there.
For ERBF, \texttt{n\_rbf}~=~\texttt{'auto'} selects the adaptive heuristic of \cref{eq:auto-k}, and the width optimisation runs for a fixed 30 iterations in gradient-based mode (\texttt{width\_optim\_iters}~=~30).
ChebyPoly constructs the Vandermonde matrix using \texttt{numpy.polynomial.chebyshev}, mapping inputs to $[-1,1]$ by min-max scaling and clipping out-of-range values (\texttt{clip\_input}~=~True) for numerical stability.
ChebyTree fits a ridge regression of the tuned degree at each leaf.

\subsection{Code availability and installation}
\label{sec:installation}

The model packages and benchmark code are publicly available under the MIT licence.

\subsubsection{Model packages} Both packages are on PyPI and can be installed with:
\begin{verbatim}
pip install erbf poly-basis-ml
\end{verbatim}
Source repositories: 
\begin{itemize}
  \item \url{https://github.com/gerberl/erbf}
  \item \url{https://github.com/gerberl/poly-basis-ml}
\end{itemize}

\subsubsection{Benchmark code}
The full benchmark--including data loaders, tuning pipeline, analysis scripts, and pre-computed results--is available at:

\url{https://github.com/gerberl/poly-erbf-benchmark}

\vspace{0.5cm}
\noindent To reproduce the experiments:

\begin{verbatim}
git clone https://github.com/gerberl/poly-erbf-benchmark.git
cd poly-erbf-benchmark
micromamba create -n polyerbfbench -f environment.yml
micromamba activate polyerbfbench
pip install erbf poly-basis-ml
python scripts/run_benchmark.py \
    --max-features 50 --max-samples 50000
\end{verbatim}
The repository README provides full instructions, including separate commands for Benchmark~A (all models, preprocessed data) and Benchmark~B (scalable models, full-scale data).

\subsection{Software versions}
\label{sec:software-versions}

All experiments were run under Python~3.12 using the \texttt{micromamba} package manager.
\Cref{tab:software-versions} lists the key packages and their versions.
\mn{tabpfn} uses the v2.5 model weights distributed with \texttt{tabpfn} package version~6.3.1 and run in local GPU mode on a 24\,GB TITAN RTX.

\begin{table}[h]
\centering
\caption{Software versions used in all experiments.}
\label{tab:software-versions}
\smallskip
\footnotesize
\begin{tabular}{lll}
\toprule
Package & Version & Role \\
\midrule
\multicolumn{3}{l}{\emph{Model implementations}} \\
\texttt{tabpfn} & 6.3.1 (v2.5 weights) & TabPFN regressor \\
\texttt{xgboost} & 3.1.3 & XGBoost \\
\texttt{scikit-learn} & 1.6.1 & Ridge, Decision Tree, Random Forest \\
\texttt{erbf} & 0.1.0 & ERBF regressor \\
\texttt{poly\_basis\_ml} & 0.2.0 & ChebyPoly, ChebyTree regressors \\
\midrule
\multicolumn{3}{l}{\emph{Infrastructure}} \\
Python & 3.12.12 & Runtime \\
\texttt{optuna} & 4.6.0 & Hyperparameter tuning \\
\texttt{numpy} & 2.3.5 & Numerical computation \\
\texttt{scipy} & 1.17.0 & Optimisation (L-BFGS-B for ERBF) \\
\texttt{pandas} & 2.3.3 & Data handling \\
\texttt{joblib} & 1.5.3 & Parallel execution and result serialisation \\
\texttt{rdkit} \citep{landrum2016rdkit} & 2025.09 & Molecular descriptor computation \\
\texttt{matplotlib} \citep{hunter2007matplotlib} & 3.10 & Figure generation \\
\texttt{seaborn} \citep{waskom2021seaborn} & 0.13 & Figure generation \\
\bottomrule
\end{tabular}
\end{table}

\section{Method-Validation Studies}
\label{sec:method-validation}

The following three analyses examine the robustness of the smooth-model design choices: ERBF sensitivity to random seeds, the numerical conditioning of the Chebyshev basis, and ERBF's sensitivity to its centre- and width-initialisation strategies.

\subsection{ERBF seed sensitivity}
\label{sec:convergence}

To assess the sensitivity of the width optimisation to initialisation, we trained ERBF with five different random seeds on five representative datasets spanning different sizes and dimensionalities.
Both ERBF sensitivity analyses use a fixed 80/20 train--test split and at most 5,000 sampled observations per dataset; for \texttt{superconduct}, the 50 features are selected by mutual information using the training data alone.
\Cref{tab:convergence} reports the mean and standard deviation of test $R^2$ and train--test gap across seeds.
Across all five datasets, the maximum standard deviation of test $R^2$ across seeds is approximately 0.0084, indicating similar predictive performance across the seeds tested.

\begin{table}[H]
\centering
\caption{ERBF sensitivity across five random seeds. Values are mean $\pm$ sample standard deviation over seeds.}
\label{tab:convergence}
\small
\begin{tabular}{@{}l cc@{}}
\toprule
\textbf{Dataset} & \textbf{$R^2$ (mean $\pm$ std)} & \textbf{Gap (mean $\pm$ std)} \\
\midrule
\texttt{california\_housing} & $0.769 \pm 0.008$ & $0.037 \pm 0.003$ \\
\texttt{concrete\_strength} & $0.845 \pm 0.008$ & $0.055 \pm 0.009$ \\
\texttt{esol} & $0.908 \pm 0.002$ & $0.020 \pm 0.001$ \\
\texttt{power\_plant} & $0.938 \pm 0.001$ & $0.010 \pm 0.001$ \\
\texttt{superconduct} & $0.848 \pm 0.003$ & $0.025 \pm 0.006$ \\
\bottomrule
\end{tabular}
\end{table}

\subsection{Chebyshev basis conditioning}
\label{sec:conditioning}

The default interaction terms are products of the scaled features, $x_i x_j = T_1(x_i)\,T_1(x_j)$, and therefore belong to the tensor-product Chebyshev basis.
This basis is orthogonal with respect to the product Chebyshev measure, but the columns of a design matrix evaluated on tabular data are not generally orthogonal, particularly when features are correlated.
The higher-complexity setting additionally appends $T_2(x_i x_j)$, which is not itself a tensor-product basis element.
We therefore assess empirical conditioning separately below.

To compare the conditioning of the Chebyshev and monomial bases, we computed the condition number of the design matrix for 21 datasets spanning all four strata, under four configurations: Chebyshev main effects, Chebyshev with interactions, monomial main effects, and monomial with interactions.
All experiments used degree~5 with the top 10 features selected by mutual information, and datasets with $n > 5{,}000$ were subsampled for SVD tractability.
Each design matrix included a single intercept column. Six datasets remained numerically singular in all four configurations; their ratios are omitted from the summary statistics.
\Cref{tab:conditioning} reports the ratio of monomial to Chebyshev condition number per dataset.

\begin{table}[H]
\centering
\caption{Ratio of monomial to Chebyshev condition number (linear scale) for 21 datasets at degree~5 with top-10 MI-selected features.  Values ${>}1$ indicate Chebyshev advantage; values ${<}1$ indicate monomial advantage.  Each matrix has one intercept. NS denotes numerical singularity (condition number greater than $1/\epsilon_{\mathrm{mach}} \approx 4.5\times10^{15}$); ratios are not reported for these cases. Datasets ordered by stratum.}
\label{tab:conditioning}
\small
\begin{tabular}{@{}l l r rr@{}}
\toprule
\textbf{Dataset} & \textbf{Stratum} & $d$ & \textbf{Main effects} & \textbf{+Interactions} \\
\midrule
\texttt{concrete\_strength}           & S1 &  8 & 4.80 & 0.94 \\
\texttt{power\_plant}                 & S1 &  4 & 2.83 & 1.00 \\
\texttt{airfoil\_noise}               & S1 &  5 & NS & NS \\
\texttt{energy\_efficiency\_heating}  & S1 &  8 & NS & NS \\
\texttt{friedman1}                    & S1 &  5 & 14.82 & 12.85 \\
\texttt{cpu\_act}                     & S1 & 10 & 1.47 & 1.25 \\
\texttt{elevators}                    & S1 & 10 & NS & NS \\
\texttt{pmlb\_225\_puma8NH}           & S1 &  8 & 13.81 & 12.91 \\
\midrule
\texttt{abalone}                      & S3 &  7 & 3.97 & 3.20 \\
\texttt{esol}                         & S3 &  7 & NS & NS \\
\texttt{lipophilicity}                & S3 & 10 & 0.86 & 0.95 \\
\texttt{qsar\_fish\_toxicity}         & S3 &  6 & NS & NS \\
\texttt{superconduct}                 & S3 & 10 & 2.14 & 0.63 \\
\texttt{sulfur}                       & S3 &  6 & 3.50 & 0.89 \\
\midrule
\texttt{wine\_quality}                & S2 & 10 & 0.97 & 1.01 \\
\texttt{analcatdata\_supreme}         & S2 &  7 & NS & NS \\
\texttt{pol}                          & S2 & 10 & 1.95 & 0.96 \\
\midrule
\texttt{california\_housing}          & S4 &  8 & 1.05 & 1.01 \\
\texttt{diamonds}                     & S4 &  6 & 0.67 & 0.71 \\
\texttt{house\_sales}                 & S4 & 10 & 0.80 & 0.84 \\
\texttt{MiamiHousing2016}             & S4 & 10 & 1.20 & 1.07 \\
\bottomrule
\end{tabular}
\end{table}

Among the 15 datasets without numerical singularity, the Chebyshev basis is better-conditioned on 11 for main effects, with a geometric mean ratio of 2.2 and a median ratio of 2.0. With interactions, it is better-conditioned on seven, with a geometric mean ratio of 1.4 and a median ratio of approximately 1.0. These results support a conditioning advantage for main effects on most of these datasets, but no consistent advantage when interactions are included.

\subsection{ERBF initialisation sensitivity}
\label{sec:lipschitz-sensitivity}

To assess ERBF's sensitivity to the choice of initialisation strategy, we evaluated all four combinations of centre placement (\texttt{lipschitz}, \texttt{kmeans}) and width initialisation (\texttt{local\_ridge}, \texttt{local\_variance}) across five representative datasets, using three random seeds per configuration.
\Cref{tab:lipschitz-sensitivity} reports the mean test $R^2$ for each combination.
The maximum range across the four configuration means is 0.0252 (\texttt{superconduct}).
The choice of centre placement and width initialisation strategy has only a modest effect on downstream performance, indicating that the three-stage pipeline is robust to these hyperparameter choices.

\begin{table}[H]
\centering
\caption{ERBF mean test $R^2$ across all four initialisation combinations (2 centre placement strategies $\times$ 2 width initialisation strategies), averaged over 3 random seeds.  Lip = Lipschitz-guided centre placement; KM = $K$-means centre placement; Ridge = local ridge width initialisation; Var = local variance width initialisation.  Range = max minus min across the four configuration means.}
\label{tab:lipschitz-sensitivity}
\small
\setlength{\tabcolsep}{4pt}
\begin{tabular}{@{}l rrrr r@{}}
\toprule
\textbf{Dataset} & \textbf{Lip/Ridge} & \textbf{Lip/Var} & \textbf{KM/Ridge} & \textbf{KM/Var} & \textbf{Range} \\
\midrule
\texttt{california\_housing} & 0.768 & 0.763 & 0.745 & 0.751 & 0.023 \\
\texttt{concrete\_strength} & 0.849 & 0.849 & 0.871 & 0.849 & 0.022 \\
\texttt{esol} & 0.907 & 0.894 & 0.899 & 0.898 & 0.014 \\
\texttt{power\_plant} & 0.938 & 0.938 & 0.938 & 0.940 & 0.002 \\
\texttt{superconduct} & 0.847 & 0.850 & 0.825 & 0.840 & 0.025 \\
\bottomrule
\end{tabular}
\end{table}

\section{Supplementary Results}
\label{sec:supplementary-results}

\subsection{Predictive accuracy rankings}
\label{sec:appendix-accuracy}

\begin{table}[H]
\centering
\caption{Predictive accuracy--all models (55 datasets, with \mn{tabpfn}). Models ordered by mean rank on $\bar{R}^2$.  Top-2 = rank-1\,/\,rank-2 counts.  Best in \textbf{bold}, second \underline{underlined}.}
\label{tab:accuracy-all}
\small
\setlength{\tabcolsep}{4pt}
\begin{tabular}{@{}lcr c rr@{}}
\toprule
\textbf{Model} & \textbf{Top-2} & \textbf{Mean Rank} & \textbf{Rank IQR} & \textbf{$\bar{R}^2$ median} & \textbf{$\bar{R}^2$ mean $\pm$ std} \\
\midrule
\textbf{\mn{tabpfn}} & \textbf{42}\,/\,6 & \textbf{1.62} & 1--1 & \textbf{0.872} & $\mathbf{0.797 \pm 0.200}$ \\
\underline{\mn{erbf}}      & 3\,/\,9 & \underline{4.09} & 3--6 & 0.818 & $\underline{0.742 \pm 0.224}$ \\
\mn{chebytree} & 2\,/\,5 & 4.15 & 3--5 & \underline{0.831} & $0.734 \pm 0.230$ \\
\mn{xgb}       & 1\,/\,\underline{15} & 4.44 & 2--6 & 0.828 & $0.735 \pm 0.225$ \\
\mn{chebypoly} & 1\,/\,8 & 4.85 & 3--6 & 0.791 & $0.708 \pm 0.228$ \\
\mn{ebm}       & 3\,/\,5 & 5.13 & 4--6 & 0.752 & $0.698 \pm 0.222$ \\
\mn{rf}        & 3\,/\,4 & 5.33 & 4--7 & 0.771 & $0.722 \pm 0.217$ \\
\mn{dt}        & 0\,/\,1 & 7.45 & 7--9 & 0.716 & $0.668 \pm 0.241$ \\
\mn{ridge}     & 0\,/\,2 & 7.95 & 8--9 & 0.597 & $0.576 \pm 0.231$ \\
\bottomrule
\end{tabular}
\end{table}

\begin{table}[H]
\centering
\caption{Predictive accuracy--CPU-viable models (55 datasets, without \mn{tabpfn}). Models ranked by mean rank on $\bar{R}^2$.  Top-2 = rank-1\,/\,rank-2 counts.  Best in \textbf{bold}, second \underline{underlined}.}
\label{tab:accuracy-cpu}
\small
\setlength{\tabcolsep}{4pt}
\begin{tabular}{@{}lcr c rr@{}}
\toprule
\textbf{Model} & \textbf{Top-2} & \textbf{Mean Rank} & \textbf{Rank IQR} & \textbf{$\bar{R}^2$ median} & \textbf{$\bar{R}^2$ mean $\pm$ std} \\
\midrule
\textbf{\mn{erbf}} & 12\,/\,12 & \textbf{3.20} & 2--5 & 0.818 & $\mathbf{0.742 \pm 0.224}$ \\
\underline{\mn{chebytree}} & 6\,/\,\underline{13} & \underline{3.25} & 2--4 & \textbf{0.831} & $0.734 \pm 0.230$ \\
\mn{xgb}       & \textbf{16}\,/\,8 & 3.47 & 1--5 & \underline{0.828} & $\underline{0.735 \pm 0.225}$ \\
\mn{chebypoly} & 7\,/\,6 & 3.95 & 3--5 & 0.791 & $0.708 \pm 0.228$ \\
\mn{ebm}       & 6\,/\,7 & 4.24 & 3--5 & 0.752 & $0.698 \pm 0.222$ \\
\mn{rf}        & 7\,/\,4 & 4.40 & 3--6 & 0.771 & $0.722 \pm 0.217$ \\
\mn{dt}        & 0\,/\,2 & 6.51 & 6--8 & 0.716 & $0.668 \pm 0.241$ \\
\mn{ridge}     & 1\,/\,3 & 6.98 & 7--8 & 0.597 & $0.576 \pm 0.231$ \\
\bottomrule
\end{tabular}
\end{table}

\subsection{Train--test gap rankings}

\begin{table}[H]
\centering
\caption{Train--test gap rankings (55 datasets, all nine models).  Lower gap = smaller train--test gap.  Models ordered by mean gap rank.  Top-2 = rank-1\,/\,rank-2 counts.  Best in \textbf{bold}, second \underline{underlined} (among competitive models; \mn{ridge} has a smaller gap but lower predictive accuracy).}
\label{tab:gap}
\small
\setlength{\tabcolsep}{3pt}
\begin{tabular}{@{}lcr c rr@{}}
\toprule
\textbf{Model} & \textbf{Top-2} & \textbf{Mean Rank} & \textbf{Rank IQR} & \textbf{Gap median} & \textbf{Gap mean $\pm$ std} \\
\midrule
\mn{ridge}     & 41\,/\,6 & 1.80 & 1--1.5 & 0.004 & $0.011 \pm 0.015$ \\
\textbf{\mn{chebypoly}} & 3\,/\,\textbf{15} & \textbf{3.55} & 2--4.5 & \textbf{0.014} & $\mathbf{0.029 \pm 0.040}$ \\
\underline{\mn{erbf}}      & 1\,/\,7 & \underline{4.27} & 3--6 & \underline{0.018} & $0.039 \pm 0.055$ \\
\mn{ebm}       & 1\,/\,7 & 4.62 & 3.5--6 & 0.020 & $\underline{0.036 \pm 0.046}$ \\
\mn{chebytree} & 2\,/\,7 & 4.85 & 3--7 & 0.026 & $0.041 \pm 0.055$ \\
\mn{rf}        & 2\,/\,5 & 5.67 & 4--7 & 0.024 & $0.062 \pm 0.098$ \\
\mn{dt}        & 1\,/\,6 & 5.84 & 4--8 & 0.032 & $0.051 \pm 0.052$ \\
\mn{tabpfn}    & 3\,/\,2 & 6.09 & 4.5--8 & 0.047 & $0.070 \pm 0.086$ \\
\mn{xgb}       & 1\,/\,0 & 8.31 & 8--9 & 0.079 & $0.107 \pm 0.107$ \\
\bottomrule
\end{tabular}
\end{table}

\subsection{Sensitivity to the matched-accuracy threshold}
\label{sec:appendix-threshold}
The matched-accuracy gap analysis (\cref{sec:results-gap}) uses a threshold of $|\Delta\bar{R}^2| \leq 0.02$ to define ``comparable accuracy.''
\Cref{tab:threshold-sensitivity} reports the comparisons separately for smooth-basis models, the ChebyTree hybrid and EBM.
The proportion of smooth-basis wins on gap remains between 85\% and 90\% across thresholds from 0.005 to 0.10, while the number of qualifying pairs increases from 39 to 202.
The finding therefore persists across the tested definitions of matched accuracy.

\begin{table}[H]
\centering
\caption{Sensitivity of the matched-accuracy gap analysis to the threshold $|\Delta\bar{R}^2|$. Each group is compared with Random Forest and XGBoost. Smooth-basis comprises ERBF and ChebyPoly. Entries give smaller-gap wins / matched pairs (percentage).}
\label{tab:threshold-sensitivity}
\small
\begin{tabular}{@{}rccc@{}}
\toprule
\textbf{Threshold} & \textbf{Smooth-basis} & \textbf{ChebyTree} & \textbf{EBM} \\
\midrule
0.005 & 35/39 (89.7\%) & 14/19 (73.7\%) & 16/19 (84.2\%) \\
0.010 & 59/67 (88.1\%) & 21/28 (75.0\%) & 24/29 (82.8\%) \\
0.020 & 88/101 (87.1\%) & 45/57 (78.9\%) & 37/43 (86.0\%) \\
0.050 & 137/160 (85.6\%) & 63/82 (76.8\%) & 71/82 (86.6\%) \\
0.100 & 172/202 (85.1\%) & 78/104 (75.0\%) & 80/93 (86.0\%) \\
\bottomrule
\end{tabular}
\end{table}

\subsection{Prediction stability: normalisation and rankings}
\label{sec:appendix-stability}
Changes in predictions after retraining are studied as \emph{predictive churn} \citep{milanifard2016churn, bahri2021churn}; resampling-based assessments also examine variation in individual predictions \citep{riley2023stability}.
The normalised prediction SD used in \cref{sec:results-stability} divides by the 1st--99th-percentile training-target span, a range measure insensitive to isolated extreme values. The measure is scale-free across datasets and unchanged by a common additive shift of targets and predictions.
Models rank identically within each dataset under this normaliser, the interquartile range, the standard deviation or the full range of the targets, since the denominator is common to all models on that dataset.
The pooled gap--stability association remains positive under alternative normalisers ($\rho = 0.80$ for the interquartile range and $0.82$ for the standard deviation, compared with $0.78$ for the interpercentile span).
With the interpercentile span, the mean within-dataset correlation is $0.58$, and the median per-model correlation across datasets is $0.76$.

Ranking each competitive model by how often it is the more stable member across its matched comparisons places \mn{chebypoly} first ($0.84$) and \mn{rf} second ($0.73$), ahead of \mn{chebytree} ($0.60$), with \mn{erbf} ($0.31$) and \mn{xgb} ($0.15$) last.

Applying the Friedman/Nemenyi procedure to per-dataset stability ranks gives a similar ordering.
Among the seven CPU-viable models (excluding \mn{ridge}), \mn{chebypoly} (mean rank $2.04$) and \mn{rf} ($2.33$) are jointly most stable -- indistinguishable from each other (difference below the critical difference, CD $= 1.21$) yet significantly more stable than the rest, and one of the two is the most stable model on 43 of 55 datasets (Friedman $p < 10^{-28}$).
Two of the significant separations fall \emph{within} a family: \mn{rf} over \mn{xgb} (rank difference $3.0$) and \mn{chebypoly} over \mn{erbf} ($2.4$).
As with the matched-accuracy counts, these are descriptive summaries under this benchmark's configurations, not a formal complexity-controlled comparison.

\subsection{Computational cost details}

\begin{table}[H]
\centering
\caption{Computational cost, averaged across 55 datasets (54 for \mn{tabpfn}). Tune/setup and train times are totals across the five outer folds; tune/setup includes hyperparameter search, model initialisation and outer-fold preprocessing. \mn{tabpfn} has no hyperparameter search, so its tune/setup entry is overhead only. Prediction cost is the summed time for predicting on training and test rows, divided by the total number of rows predicted across folds and expressed per 1\,000 rows; it measures batch prediction cost. Models ordered by tune/setup time. Fastest competitive CPU model in \textbf{bold}, second \underline{underlined} (tune/setup).}
\label{tab:timing}
\small
\begin{tabular}{@{}l rrr@{}}
\toprule
\textbf{Model} & \textbf{Tune/setup (s)} & \textbf{Train (s)} & \textbf{Predict (ms/1K)} \\
\midrule
\mn{tabpfn} & 2.4 & 4.19 & 1253.15 \\
\mn{ridge} & 19.2 & 0.01 & 0.19 \\
\mn{dt} & 28.0 & 0.28 & 0.32 \\
\textbf{\mn{chebypoly}} & \textbf{38.2} & 0.36 & 3.48 \\
\underline{\mn{chebytree}} & \underline{59.5} & 0.50 & 7.27 \\
\mn{xgb} & 189.5 & 1.15 & 3.35 \\
\mn{ebm} & 431.5 & 10.45 & 0.46 \\
\mn{erbf} & 799.7 & 17.82 & 2.12 \\
\mn{rf} & 1395.2 & 42.08 & 51.15 \\
\bottomrule
\end{tabular}
\end{table}

\Cref{tab:timing-examples} illustrates tuning/setup and training totals on three datasets spanning the size range, using the same five-fold convention.

\begin{table}[H]
\centering
\caption{Tuning/setup and training times (seconds, summed across five outer folds) on three datasets spanning the size range, for the CPU models.}
\label{tab:timing-examples}
\small
\resizebox{\textwidth}{!}{\begin{tabular}{@{}l l rr rrrrrrrr@{}}
\toprule
\textbf{Dataset} & & \textbf{$n$} & \textbf{$d$} & \textbf{ridge} & \textbf{dt} & \textbf{rf} & \textbf{xgb} & \textbf{ebm} & \textbf{erbf} & \textbf{chebypoly} & \textbf{chebytree} \\
\midrule
\texttt{concrete\_strength} & tune/setup & 1\,030 & 8 & 5.17 & 5.64 & 254.49 & 60.93 & 251.17 & 66.81 & 7.29 & 24.88 \\
 & train &  &  & 0.01 & 0.02 & 6.46 & 0.47 & 5.95 & 1.04 & 0.03 & 0.10 \\
\addlinespace
\texttt{california\_housing} & tune/setup & 20\,640 & 8 & 6.85 & 19.60 & 2309.51 & 243.11 & 502.44 & 768.85 & 19.55 & 70.67 \\
 & train &  &  & 0.01 & 0.41 & 48.26 & 1.22 & 12.28 & 18.12 & 0.23 & 1.10 \\
\addlinespace
\texttt{superconduct} & tune/setup & 21\,263 & 79 & 134.89 & 177.66 & 7318.43 & 1093.18 & 2693.85 & 3813.78 & 220.53 & 216.37 \\
 & train &  &  & 0.04 & 1.87 & 207.46 & 8.99 & 63.93 & 84.55 & 3.08 & 2.41 \\
\bottomrule
\end{tabular}}
\end{table}

\subsection{Per-dataset results}
\label{sec:per-dataset-results}

\Cref{tab:per-dataset-rank} gives the per-dataset ranks over the eight CPU-viable models (rank~1 = best on each dataset); these average to the mean ranks reported in \cref{sec:results-accuracy}.
\Cref{tab:per-dataset-r2adj} reports the corresponding per-dataset adjusted $R^2$ for all nine models (overall best per dataset in \textbf{bold}; best CPU-viable model \underline{underlined}).

{\scriptsize
\setlength{\tabcolsep}{1.5pt}
\renewcommand{\mn}[1]{\texttt{#1}}\begin{longtable}{@{}lrrrrrrrr@{}}
\caption{Per-dataset rank of the eight CPU-viable models by adjusted $R^2$ (rank~1 = best on that dataset; ties share the lower rank).  Best per dataset in \textbf{bold}.  $^\dagger$Ordinal (discrete integer) target.  TabPFN is excluded (analysed separately).  Mean ranks per model are reported in the main text.}
\label{tab:per-dataset-rank} \\
\toprule
\textbf{Dataset} & \mn{erbf} & \mn{chebytree} & \mn{xgb} & \mn{ebm} & \mn{chebypoly} & \mn{rf} & \mn{dt} & \mn{ridge} \\
\midrule
\endfirsthead
\toprule
\textbf{Dataset} & \mn{erbf} & \mn{chebytree} & \mn{xgb} & \mn{ebm} & \mn{chebypoly} & \mn{rf} & \mn{dt} & \mn{ridge} \\
\midrule
\endhead
\midrule \multicolumn{9}{r}{\emph{Continued on next page}} \\
\endfoot
\bottomrule
\endlastfoot
\multicolumn{9}{@{}l}{\emph{S1: Engineering/Simulation}} \\
\addlinespace[2pt]
\texttt{Ailerons} & 5 & 2 & 3 & 4 & \textbf{1} & 7 & 8 & 6 \\
\texttt{airfoil\_noise} & 2 & 3 & \textbf{1} & 7 & 5 & 4 & 6 & 8 \\
\texttt{concrete\_strength} & 4 & 6 & \textbf{1} & 2 & 3 & 5 & 7 & 8 \\
\texttt{cpu\_act} & 2 & 3 & 4 & 5 & \textbf{1} & 6 & 7 & 8 \\
\texttt{elevators} & 5 & 2 & 3 & 4 & \textbf{1} & 7 & 8 & 6 \\
\texttt{energy\_efficiency\_heating} & \textbf{1} & 2 & 4 & 7 & 3 & 6 & 5 & 8 \\
\texttt{feynman\_gaussian} & 2 & \textbf{1} & 6 & 5 & 7 & 3 & 4 & 8 \\
\texttt{feynman\_wave\_interference} & \textbf{1} & 5 & 2 & 7 & 3 & 4 & 6 & 8 \\
\texttt{friedman1} & \textbf{1} & 3 & 2 & 5 & 4 & 6 & 7 & 8 \\
\texttt{friedman1\_d100} & \textbf{1} & 3 & 2 & 5 & 4 & 6 & 8 & 7 \\
\texttt{pmlb\_215\_2dplanes} & 4 & \textbf{1} & 5 & 8 & 3 & 6 & 2 & 7 \\
\texttt{pmlb\_225\_puma8NH} & \textbf{1} & 3 & 4 & 7 & 6 & 2 & 5 & 8 \\
\texttt{power\_plant} & 3 & 5 & \textbf{1} & 2 & 6 & 4 & 7 & 8 \\
\addlinespace[4pt]
\multicolumn{9}{@{}l}{\emph{S2: Behavioural/Social}} \\
\addlinespace[2pt]
\texttt{Bike\_Sharing\_Demand} & 3 & 2 & \textbf{1} & 7 & 6 & 4 & 5 & 8 \\
\texttt{analcatdata\_supreme} & 6 & 4 & 5 & \textbf{1} & 2 & 7 & 3 & 8 \\
\texttt{food\_delivery\_time} & 7 & 3 & \textbf{1} & 2 & 4 & 5 & 6 & 8 \\
\texttt{pmlb\_1028\_SWD$^\dagger$} & 5 & 2 & 7 & \textbf{1} & 4 & 3 & 8 & 6 \\
\texttt{pmlb\_1029\_LEV$^\dagger$} & 3 & 5 & 7 & 4 & \textbf{1} & 6 & 8 & 2 \\
\texttt{pmlb\_1030\_ERA$^\dagger$} & 4 & \textbf{1} & 8 & 2 & 6 & 5 & 7 & 3 \\
\texttt{pmlb\_4544\_GeographicalOrigi...} & 6 & 3 & 7 & 5 & \textbf{1} & 4 & 8 & 2 \\
\texttt{pol$^\dagger$} & 3 & 2 & \textbf{1} & 7 & 6 & 4 & 5 & 8 \\
\texttt{student\_performance$^\dagger$} & 6 & 3 & 7 & 4 & 2 & \textbf{1} & 8 & 5 \\
\texttt{wine\_quality$^\dagger$} & 3 & 6 & \textbf{1} & 5 & 4 & 2 & 7 & 8 \\
\addlinespace[4pt]
\multicolumn{9}{@{}l}{\emph{S3: Physics/Chemistry/Life Sciences}} \\
\addlinespace[2pt]
\texttt{abalone} & 3 & \textbf{1} & 6 & 5 & 2 & 4 & 8 & 7 \\
\texttt{diabetes} & 6 & 3 & 7 & 4 & \textbf{1} & 5 & 8 & 2 \\
\texttt{esol} & \textbf{1} & 7 & 4 & 5 & 3 & 2 & 6 & 8 \\
\texttt{freesolv} & 3 & 4 & 5 & \textbf{1} & 2 & 6 & 8 & 7 \\
\texttt{lipophilicity} & 3 & 6 & 2 & 4 & 5 & \textbf{1} & 7 & 8 \\
\texttt{particulate-matter-ukair-2017} & 5 & 3 & 4 & \textbf{1} & 2 & 7 & 8 & 6 \\
\texttt{physiochemical\_protein} & 2 & 3 & \textbf{1} & 6 & 5 & 4 & 7 & 8 \\
\texttt{pmlb\_503\_wind} & 2 & 3 & 5 & 4 & \textbf{1} & 6 & 8 & 7 \\
\texttt{pmlb\_522\_pm10} & \textbf{1} & 8 & 5 & 2 & 4 & 3 & 6 & 7 \\
\texttt{pmlb\_529\_pollen} & 4 & 2 & 6 & 5 & 3 & 7 & 8 & \textbf{1} \\
\texttt{pmlb\_547\_no2} & 2 & 4 & 6 & 3 & 5 & \textbf{1} & 8 & 7 \\
\texttt{qm7} & \textbf{1} & 6 & 2 & 5 & 3 & 4 & 8 & 7 \\
\texttt{qsar\_fish\_toxicity} & 2 & 4 & 7 & 3 & 5 & \textbf{1} & 8 & 6 \\
\texttt{qsar\_tid\_11} & 3 & 2 & \textbf{1} & 8 & 5 & 4 & 6 & 7 \\
\texttt{sulfur} & 3 & 2 & \textbf{1} & 5 & 4 & 6 & 7 & 8 \\
\texttt{superconduct} & 5 & 2 & \textbf{1} & 3 & 7 & 4 & 6 & 8 \\
\addlinespace[4pt]
\multicolumn{9}{@{}l}{\emph{S4: Economic/Pricing}} \\
\addlinespace[2pt]
\texttt{Allstate\_Claims\_Severity} & 2 & \textbf{1} & 3 & 5 & 4 & 7 & 8 & 6 \\
\texttt{Brazilian\_houses} & 6 & 2 & \textbf{1} & 4 & 7 & 3 & 5 & 8 \\
\texttt{MiamiHousing2016} & \textbf{1} & 4 & 2 & 5 & 3 & 6 & 7 & 8 \\
\texttt{california\_housing} & 2 & 4 & \textbf{1} & 3 & 6 & 5 & 7 & 8 \\
\texttt{diamonds} & 2 & 5 & \textbf{1} & 3 & 6 & 4 & 7 & 8 \\
\texttt{fiat\_500\_price} & 5 & 4 & 7 & 2 & 3 & \textbf{1} & 8 & 6 \\
\texttt{healthcare\_insurance} & 5 & \textbf{1} & 3 & 7 & 6 & 2 & 4 & 8 \\
\texttt{house\_16H} & 5 & 4 & 2 & \textbf{1} & 3 & 6 & 7 & 8 \\
\texttt{house\_sales} & 2 & 3 & \textbf{1} & 4 & 5 & 6 & 7 & 8 \\
\texttt{medical\_charges} & \textbf{1} & 2 & 5 & 4 & 3 & 7 & 6 & 8 \\
\texttt{nyc-taxi-green-dec-2016} & 5 & 2 & 4 & \textbf{1} & 7 & 6 & 3 & 8 \\
\texttt{pmlb\_218\_house\_8L} & 2 & 3 & \textbf{1} & 6 & 4 & 5 & 7 & 8 \\
\texttt{power\_grid\_stability} & \textbf{1} & 3 & 2 & 6 & 4 & 5 & 7 & 8 \\
\texttt{synthetic\_multithreshold} & 6 & 5 & 3 & 2 & 7 & \textbf{1} & 4 & 8 \\
\texttt{synthetic\_piecewise} & \textbf{1} & 3 & 4 & 6 & 2 & 5 & 7 & 8 \\
\texttt{synthetic\_step} & 6 & 3 & 5 & 4 & 7 & \textbf{1} & 2 & 8 \\
\end{longtable} }

{\scriptsize
\setlength{\tabcolsep}{1.5pt}
\renewcommand{\mn}[1]{\texttt{#1}}\begin{longtable}{@{}lrrrrrrrrr@{}}
\caption{Per-dataset adjusted $R^2$ for all nine models, grouped by stratum.  Overall best per dataset in \textbf{bold}; best CPU-viable model \underline{underlined} (shown only when different from overall best).  $^\dagger$Ordinal (discrete integer) target.}
\label{tab:per-dataset-r2adj} \\
\toprule
\textbf{Dataset} & \mn{tabpfn} & \mn{erbf} & \mn{chebytree} & \mn{xgb} & \mn{ebm} & \mn{chebypoly} & \mn{rf} & \mn{dt} & \mn{ridge} \\
\midrule
\endfirsthead
\toprule
\textbf{Dataset} & \mn{tabpfn} & \mn{erbf} & \mn{chebytree} & \mn{xgb} & \mn{ebm} & \mn{chebypoly} & \mn{rf} & \mn{dt} & \mn{ridge} \\
\midrule
\endhead
\midrule \multicolumn{10}{r}{\emph{Continued on next page}} \\
\endfoot
\bottomrule
\endlastfoot
\multicolumn{10}{@{}l}{\emph{S1: Engineering/Simulation}} \\
\addlinespace[2pt]
\texttt{Ailerons} & \textbf{0.869} & 0.818 & 0.841 & 0.826 & 0.824 & \underline{0.844} & 0.806 & 0.758 & 0.818 \\
\texttt{airfoil\_noise} & \textbf{0.977} & 0.921 & 0.831 & \underline{0.923} & 0.557 & 0.728 & 0.828 & 0.725 & 0.503 \\
\texttt{concrete\_strength} & \textbf{0.945} & 0.885 & 0.865 & \underline{0.917} & 0.907 & 0.892 & 0.880 & 0.809 & 0.597 \\
\texttt{cpu\_act} & \textbf{0.988} & 0.983 & 0.982 & 0.982 & 0.981 & \underline{0.983} & 0.976 & 0.968 & 0.736 \\
\texttt{elevators} & \textbf{0.929} & 0.862 & 0.886 & 0.873 & 0.873 & \underline{0.890} & 0.689 & 0.594 & 0.812 \\
\texttt{energy\_efficiency\_heating} & 0.998 & \textbf{0.998} & 0.997 & 0.996 & 0.989 & 0.997 & 0.995 & 0.996 & 0.913 \\
\texttt{feynman\_gaussian} & \textbf{0.999} & 0.969 & \underline{0.991} & 0.490 & 0.500 & 0.381 & 0.945 & 0.939 & 0.068 \\
\texttt{feynman\_wave\_interference} & \textbf{0.999} & \underline{0.965} & 0.883 & 0.935 & 0.846 & 0.909 & 0.888 & 0.873 & 0.841 \\
\texttt{friedman1} & \textbf{0.999} & \underline{0.998} & 0.948 & 0.961 & 0.920 & 0.923 & 0.883 & 0.751 & 0.741 \\
\texttt{friedman1\_d100} & \textbf{0.958} & \underline{0.942} & 0.889 & 0.904 & 0.870 & 0.875 & 0.846 & 0.716 & 0.736 \\
\texttt{pmlb\_215\_2dplanes} & 0.948 & 0.948 & \textbf{0.948} & 0.947 & 0.705 & 0.948 & 0.947 & 0.948 & 0.705 \\
\texttt{pmlb\_225\_puma8NH} & \textbf{0.687} & \underline{0.686} & 0.671 & 0.656 & 0.430 & 0.431 & 0.676 & 0.649 & 0.368 \\
\texttt{power\_plant} & \textbf{0.967} & 0.948 & 0.944 & \underline{0.961} & 0.952 & 0.941 & 0.947 & 0.939 & 0.928 \\
\addlinespace[4pt]
\multicolumn{10}{@{}l}{\emph{S2: Behavioural/Social}} \\
\addlinespace[2pt]
\texttt{Bike\_Sharing\_Demand} & \textbf{0.718} & 0.671 & 0.682 & \underline{0.689} & 0.631 & 0.644 & 0.668 & 0.654 & 0.333 \\
\texttt{analcatdata\_supreme} & 0.977 & 0.977 & 0.978 & 0.978 & \textbf{0.978} & 0.978 & 0.972 & 0.978 & 0.429 \\
\texttt{food\_delivery\_time} & --- & 0.252 & 0.335 & \textbf{0.346} & 0.336 & 0.331 & 0.284 & 0.278 & 0.181 \\
\texttt{pmlb\_1028\_SWD$^\dagger$} & 0.412 & 0.405 & 0.412 & 0.387 & \textbf{0.415} & 0.409 & 0.409 & 0.363 & 0.388 \\
\texttt{pmlb\_1029\_LEV$^\dagger$} & 0.544 & 0.548 & 0.538 & 0.501 & 0.545 & \textbf{0.551} & 0.524 & 0.478 & 0.551 \\
\texttt{pmlb\_1030\_ERA$^\dagger$} & 0.352 & 0.344 & \textbf{0.365} & 0.325 & 0.356 & 0.340 & 0.342 & 0.334 & 0.351 \\
\texttt{pmlb\_4544\_GeographicalOrigi...} & \textbf{0.799} & 0.697 & 0.731 & 0.675 & 0.701 & \underline{0.774} & 0.705 & 0.521 & 0.774 \\
\texttt{pol$^\dagger$} & \textbf{0.986} & 0.962 & 0.963 & \underline{0.977} & 0.733 & 0.868 & 0.949 & 0.925 & 0.462 \\
\texttt{student\_performance$^\dagger$} & 0.275 & 0.212 & 0.232 & 0.195 & 0.231 & 0.239 & \textbf{0.275} & 0.147 & 0.228 \\
\texttt{wine\_quality$^\dagger$} & \textbf{0.514} & 0.387 & 0.335 & \underline{0.439} & 0.353 & 0.363 & 0.391 & 0.301 & 0.287 \\
\addlinespace[4pt]
\multicolumn{10}{@{}l}{\emph{S3: Physics/Chemistry/Life Sciences}} \\
\addlinespace[2pt]
\texttt{abalone} & \textbf{0.586} & 0.556 & \underline{0.564} & 0.521 & 0.526 & 0.559 & 0.548 & 0.481 & 0.514 \\
\texttt{diabetes} & \textbf{0.473} & 0.432 & 0.460 & 0.409 & 0.455 & \underline{0.464} & 0.435 & 0.296 & 0.464 \\
\texttt{esol} & \textbf{0.916} & \underline{0.896} & 0.847 & 0.881 & 0.858 & 0.893 & 0.894 & 0.852 & 0.827 \\
\texttt{freesolv} & \textbf{0.935} & 0.885 & 0.884 & 0.882 & \underline{0.898} & 0.886 & 0.877 & 0.782 & 0.823 \\
\texttt{lipophilicity} & \textbf{0.592} & 0.387 & 0.329 & 0.390 & 0.343 & 0.331 & \underline{0.397} & 0.293 & 0.247 \\
\texttt{particulate-matter-ukair-2017} & 0.744 & 0.740 & 0.746 & 0.741 & \textbf{0.752} & 0.750 & 0.726 & 0.718 & 0.733 \\
\texttt{physiochemical\_protein} & \textbf{0.768} & 0.505 & 0.505 & \underline{0.615} & 0.378 & 0.404 & 0.424 & 0.359 & 0.281 \\
\texttt{pmlb\_503\_wind} & \textbf{0.815} & 0.791 & 0.785 & 0.768 & 0.784 & \underline{0.799} & 0.760 & 0.706 & 0.758 \\
\texttt{pmlb\_522\_pm10} & 0.398 & \textbf{0.398} & 0.094 & 0.296 & 0.377 & 0.349 & 0.365 & 0.132 & 0.123 \\
\texttt{pmlb\_529\_pollen} & \textbf{0.794} & 0.791 & 0.792 & 0.740 & 0.782 & 0.791 & 0.710 & 0.639 & \underline{0.792} \\
\texttt{pmlb\_547\_no2} & \textbf{0.600} & 0.588 & 0.543 & 0.519 & 0.555 & 0.539 & \underline{0.599} & 0.469 & 0.476 \\
\texttt{qm7} & \textbf{0.810} & \underline{0.785} & 0.768 & 0.779 & 0.771 & 0.773 & 0.771 & 0.737 & 0.746 \\
\texttt{qsar\_fish\_toxicity} & \textbf{0.639} & 0.611 & 0.564 & 0.552 & 0.584 & 0.559 & \underline{0.615} & 0.498 & 0.556 \\
\texttt{qsar\_tid\_11} & \textbf{0.650} & 0.526 & 0.535 & \underline{0.607} & 0.315 & 0.398 & 0.468 & 0.393 & 0.317 \\
\texttt{sulfur} & \textbf{0.920} & 0.779 & 0.791 & \underline{0.828} & 0.748 & 0.764 & 0.731 & 0.716 & 0.393 \\
\texttt{superconduct} & \textbf{0.931} & 0.848 & 0.890 & \underline{0.909} & 0.888 & 0.819 & 0.856 & 0.819 & 0.675 \\
\addlinespace[4pt]
\multicolumn{10}{@{}l}{\emph{S4: Economic/Pricing}} \\
\addlinespace[2pt]
\texttt{Allstate\_Claims\_Severity} & \textbf{0.538} & 0.494 & \underline{0.500} & 0.480 & 0.459 & 0.470 & 0.438 & 0.401 & 0.450 \\
\texttt{Brazilian\_houses} & \textbf{0.996} & 0.972 & 0.982 & \underline{0.987} & 0.978 & 0.966 & 0.979 & 0.973 & 0.836 \\
\texttt{MiamiHousing2016} & \textbf{0.944} & \underline{0.918} & 0.897 & 0.917 & 0.895 & 0.909 & 0.864 & 0.802 & 0.717 \\
\texttt{california\_housing} & \textbf{0.875} & 0.792 & 0.751 & \underline{0.829} & 0.772 & 0.730 & 0.739 & 0.667 & 0.601 \\
\texttt{diamonds} & \textbf{0.947} & 0.945 & 0.943 & \underline{0.945} & 0.945 & 0.943 & 0.944 & 0.942 & 0.925 \\
\texttt{fiat\_500\_price} & \textbf{0.857} & 0.841 & 0.844 & 0.834 & 0.850 & 0.846 & \underline{0.851} & 0.823 & 0.840 \\
\texttt{healthcare\_insurance} & \textbf{0.856} & 0.839 & \underline{0.856} & 0.845 & 0.737 & 0.833 & 0.855 & 0.840 & 0.733 \\
\texttt{house\_16H} & \textbf{0.583} & 0.491 & 0.502 & 0.514 & \underline{0.515} & 0.508 & 0.485 & 0.422 & 0.235 \\
\texttt{house\_sales} & \textbf{0.903} & 0.870 & 0.852 & \underline{0.876} & 0.847 & 0.846 & 0.838 & 0.790 & 0.746 \\
\texttt{medical\_charges} & 0.980 & \textbf{0.980} & 0.980 & 0.979 & 0.980 & 0.980 & 0.977 & 0.978 & 0.827 \\
\texttt{nyc-taxi-green-dec-2016} & \textbf{0.578} & 0.459 & 0.530 & 0.513 & \underline{0.568} & 0.393 & 0.440 & 0.517 & 0.307 \\
\texttt{pmlb\_218\_house\_8L} & \textbf{0.724} & 0.661 & 0.657 & \underline{0.672} & 0.619 & 0.638 & 0.621 & 0.575 & 0.380 \\
\texttt{power\_grid\_stability} & \textbf{0.987} & \underline{0.951} & 0.896 & 0.926 & 0.784 & 0.895 & 0.815 & 0.689 & 0.645 \\
\texttt{synthetic\_multithreshold} & 0.965 & 0.888 & 0.931 & 0.959 & 0.963 & 0.858 & \textbf{0.966} & 0.958 & 0.569 \\
\texttt{synthetic\_piecewise} & \textbf{0.963} & \underline{0.960} & 0.952 & 0.935 & 0.920 & 0.955 & 0.935 & 0.902 & 0.791 \\
\texttt{synthetic\_step} & 0.939 & 0.877 & 0.932 & 0.922 & 0.923 & 0.844 & \textbf{0.939} & 0.938 & 0.579 \\
\end{longtable} }

\subsection{ERBF performance on discrete targets}
\label{sec:erbf-ordinal}

ERBF's mean rank drops from 2.88 on continuous-target datasets to 4.23 on the thirteen datasets with non-continuous targets (and 4.00 on the nine datasets with discrete or ordinal targets shown in \cref{tab:erbf-ordinal-full}).
However, with only thirteen non-continuous datasets in our benchmark, this rank difference should be interpreted cautiously: the sample is too small for the shift to be statistically reliable.
We present the analysis below as exploratory rather than confirmatory.

\Cref{tab:erbf-ordinal-full} compares ERBF and XGBoost with the best of the remaining six CPU-viable models on the nine datasets using adjusted $R^2$ with discrete or ordinal targets, ordered by the number of unique target values.
ERBF's performance is mixed: it outperforms XGBoost on several datasets (e.g., \texttt{student\_performance} $+0.02$, \texttt{abalone} $+0.03$, \texttt{pmlb\_1029\_LEV} $+0.05$) but underperforms it on others (\texttt{wine\_quality} $-0.05$, \texttt{pol} $-0.01$).
The pattern does not reduce to a simple threshold on the number of distinct target values, and the small sample precludes identifying a reliable predictor of when ERBF will underperform.
Plausible contributing factors include the combination of few target levels, categorical input features (which produce target-encoded representations with limited gradient information), and small sample size; these are confounded across the nine datasets, and separating their individual contributions is a natural direction for future work.
For instance, \texttt{student\_performance} ($n = 649$, 17 unique $y$, 17 categorical features) is challenging for all models ($\bar{R}^2 = 0.15$--$0.27$), with ERBF's smooth surface providing no advantage over tree-based alternatives.

\begin{table}[H]
\centering
\caption{Adjusted $R^2$ on the nine datasets with discrete or ordinal targets, ordered by number of unique target values.  ERBF and XGBoost are shown alongside the best-performing model among the remaining six (named in the rightmost column).  All results use Optuna-tuned hyperparameters from the main benchmark.  \texttt{Bike\_Sharing\_Demand} is an integer count with many distinct values (869); it is the highest-cardinality of the thirteen non-continuous datasets (\cref{tab:target-type}) and is included here as a contrast to the low-resolution ordinal and count targets.}
\label{tab:erbf-ordinal-full}
\small
\setlength{\tabcolsep}{1.5pt}
\begin{tabular}{@{}l rr rrr l@{}}
\toprule
\textbf{Dataset} & \textbf{Unique $y$} & $n$ & \textbf{ERBF} & \textbf{XGB} & \textbf{Best other} & \textbf{Model} \\
\midrule
\texttt{pmlb\_1028\_SWD}      &   4 & 1\,000 & 0.405 & 0.387 & \textbf{0.415} & ebm \\
\texttt{pmlb\_1029\_LEV}      &   5 & 1\,000 & 0.548 & 0.501 & \textbf{0.551} & chebypoly \\
\texttt{wine\_quality}        &   7 & 6\,497 & 0.387 & \textbf{0.439} & 0.391 & rf \\
\texttt{pmlb\_1030\_ERA}      &   9 & 1\,000 & 0.344 & 0.325 & \textbf{0.365} & chebytree \\
\texttt{analcatdata\_supreme} &  10 & 4\,052 & 0.977 & 0.978 & \textbf{0.978} & ebm \\
\texttt{pol}                  &  11 & 15\,000 & 0.962 & \textbf{0.977} & 0.963 & chebytree \\
\texttt{student\_performance} &  17 &    649 & 0.212 & 0.195 & \textbf{0.275} & rf \\
\texttt{abalone}              &  28 & 4\,177 & 0.556 & 0.521 & \textbf{0.564} & chebytree \\
\texttt{Bike\_Sharing\_Demand} & 869 & 17\,379 & 0.671 & \textbf{0.689} & 0.682 & chebytree \\
\bottomrule
\end{tabular}
\end{table}

To examine the effect of target discretisation, we constructed a synthetic experiment with 1,000 observations.
A smooth function ($\sin(2\pi x_1) + \cos(3\pi x_2) + \varepsilon$, $\varepsilon \sim \mathcal{N}(0, 0.01)$, with three noise features) was rounded to between 3 and 100 equally spaced levels spanning the observed target range, alongside an unrounded comparison.
The inputs and five cross-validation folds were held fixed across conditions; each model was evaluated against the corresponding rounded or continuous target.
\Cref{tab:erbf-ordinal-synth} reports mean test $R^2$ for ERBF, XGBoost, and a decision tree using default hyperparameters and random seed 42.

On this synthetic problem, ERBF achieved the highest mean test $R^2$ from five target levels upwards, but fell behind both tree models at three levels.
Severe target discretisation can therefore disadvantage ERBF even when the underlying function is smooth, although this single example does not establish the cause of the mixed results on real discrete-target datasets.
The dip at 50 levels includes one seed-sensitive ERBF fit ($R^2 = 0.939$); refitting that fold with three other seeds gave $R^2 \approx 0.989$, but the original fit is retained in the table.

\begin{table}[H]
\centering
\caption{Synthetic experiment: effect of target rounding on model accuracy.  A smooth function is discretised to varying numbers of equally spaced levels.  Default hyperparameters; mean test $R^2$ over the same five folds (seed 42).}
\label{tab:erbf-ordinal-synth}
\small
\begin{tabular}{@{}r rrr@{}}
\toprule
\textbf{Levels} & \textbf{ERBF $R^2$} & \textbf{XGB $R^2$} & \textbf{DT $R^2$} \\
\midrule
3 & 0.797 & 0.855 & 0.809 \\
5 & 0.894 & 0.886 & 0.847 \\
7 & 0.943 & 0.923 & 0.878 \\
10 & 0.966 & 0.945 & 0.909 \\
15 & 0.979 & 0.960 & 0.918 \\
21 & 0.984 & 0.964 & 0.921 \\
50 & 0.977 & 0.969 & 0.932 \\
100 & 0.986 & 0.969 & 0.927 \\
Continuous & 0.987 & 0.969 & 0.929 \\
\bottomrule
\end{tabular}
\end{table}

\end{document}